\documentclass[letterpaper]{article} 
\usepackage{aaai2026}  
\usepackage{times}  
\usepackage{helvet}  
\usepackage{courier}  
\usepackage[hyphens]{url}  
\usepackage{graphicx} 
\usepackage{natbib}  
\usepackage{caption} 
\usepackage{algorithm}
\usepackage{algpseudocode} 
\usepackage{amsmath}
\usepackage{amssymb}
\usepackage{booktabs}
\usepackage{multirow} 
\usepackage{float}
\usepackage{subcaption}
\usepackage{tabularx}
\usepackage[most]{tcolorbox}
\usepackage{newfloat}
\usepackage{listings}
\DeclareCaptionStyle{ruled}{labelfont=normalfont,labelsep=colon,strut=off} 
\floatstyle{ruled}
\newfloat{listing}{tb}{lst}{}
\floatname{listing}{Listing}

\usepackage{xspace}
\newcommand{\factguard}{\textsc{FactGuard}\xspace}

\usepackage{arydshln}
\usepackage{bm} 
\usepackage{pifont} 

\usepackage{xcolor}           
\usepackage{tcolorbox}
\usepackage{amsmath}
\usepackage{enumitem}
\usepackage{makecell}

\title{\factguard: Event-Centric and Commonsense-Guided Fake News Detection}
\author{
    Jing He\textsuperscript{\rm 1}, Han Zhang\textsuperscript{\rm 1}, Yuanhui Xiao\textsuperscript{\rm 1}, Wei Guo\textsuperscript{\rm 1}, Shaowen Yao\textsuperscript{\rm 1}, Renyang Liu\textsuperscript{\rm 2,}\thanks{Corresponding author},
}

\affiliations{
    \textsuperscript{\rm 1}School of Software and AI, Yunnan University\\
    \textsuperscript{\rm 2}Institute of Data Science, National University of Singapore
    
    hejing@ynu.edu.cn,\{hanzhang, xiaoyuanhui, guoweiynu\}@stu.ynu.edu.cn, yaosw@ynu.edu.cn, ryliu@nus.edu.sg
}

\begin{document}

\maketitle

\begin{abstract}
Fake news detection methods based on writing style have achieved remarkable progress. However, as adversaries increasingly imitate the style of authentic news, the effectiveness of such approaches is gradually diminishing. Recent research has explored incorporating large language models (LLMs) to enhance fake news detection. Yet, despite their transformative potential, LLMs remain an untapped goldmine for fake news detection, with their real-world adoption hampered by shallow functionality exploration, ambiguous usability, and prohibitive inference costs.
In this paper, we propose a novel fake news detection framework, dubbed \factguard, that leverages LLMs to extract event-centric content, thereby reducing the impact of writing style on detection performance. Furthermore, our approach introduces a dynamic usability mechanism that identifies contradictions and ambiguous cases in factual reasoning, adaptively incorporating LLM advice to improve decision reliability. To ensure efficiency and practical deployment, we employ knowledge distillation to derive \factguard-D, enabling the framework to operate effectively in cold-start and resource-constrained scenarios.
Comprehensive experiments on two benchmark datasets demonstrate that our approach consistently outperforms existing methods in both robustness and accuracy, effectively addressing the challenges of style sensitivity and LLM usability in fake news detection.
\end{abstract}

\begin{links}
    \link{Code}{https://github.com/ryliu68/FACTGUARD}
\end{links}

\section{Introduction}
Social media platforms have become dominant channels for information dissemination, surpassing traditional media in both scale and societal influence. However, the lack of effective content moderation on these platforms has facilitated the rapid spread of sensational fake news, further amplified by recommendation algorithms. Consequently, platforms such as Sina-Weibo and Facebook have faced significant challenges related to misinformation propagation~\cite{weibo2021,avaaz2020facebook}. Prior studies have demonstrated that the widespread circulation of fake news during major public events can trigger social panic and disrupt governance~\cite{grinberg2019fake,zhang2024enhancing,bursztyn2020misinformation}. Given the overwhelming volume of online information, manual verification of news authenticity is infeasible, which underscores the necessity of developing automated fake news detection methods based on advanced techniques, such as deep learning, with a particular emphasis on early identification to curb the spread and societal impact of misinformation.

Early efforts in this direction have predominantly relied on insights from psychology and linguistics. In particular, a substantial body of research is rooted in psychological theories, such as the Undeutsch hypothesis~\cite{amado2015undeutsch}, emphasizing linguistic style differences between truthful and deceptive statements. Stylometric and emotion-based methods~\cite{potthast2018stylometric,rashkin2017truth,ajao2019sentiment,giachanou2019leveraging,zhu2022generalizing} have thus become mainstream approaches for identifying fake news. However, as these methods primarily capture superficial features, recent studies have shown that adversaries can effectively evade detection by imitating the writing style of authentic news. This reliance on surface-level cues makes existing systems highly vulnerable to news text writing style.

To address these limitations, the research community has increasingly explored leveraging large language models (LLMs) for fake news detection. For example, some studies employ LLMs to generate adversarial samples with diverse writing styles to enhance model robustness~\cite{wu2024fake,wang2024megafake}. Other approaches integrate multiple perspectives, such as combining style detection with commonsense reasoning~\cite{hu2024bad}, constructing multi-agent debate frameworks to aggregate different viewpoints~\cite{liu2025truth} or using LLMs to simulate different news readers to generate diverse comments~\cite{nan2024let}. Despite these methodological advances, several critical challenges remain unresolved. In particular, the effectiveness of style-based sample generation methods against previously unseen style attacks remains uncertain. LLM-based detection models, though promising, often suffer from low accuracy in few-shot and chain-of-thought reasoning scenarios~\cite{hu2024bad}, are prone to hallucinations~\cite{xu2024hallucination}. Besides, some methods focus only on correctly judged samples during training, while the correctness of such judgments remains unknown during inference~\cite{hu2024bad}, resulting in the lack of reliable mechanisms for usability assessment. Moreover, multi-agent debate frameworks and role-playing–based comment generation typically incur considerable computational and time costs, which limits their practicality in cold-start\footnote{Cold-start refers to the early-stage fake news detection setting where only the news content is available. In this setting, high inference efficiency is required.} and resource-constrained environments\footnote{Resource-constrained refers to settings where LLMs cannot be accessed or invoked, and the model can only rely on news content for inference in such cases.}~\cite{liu2025truth}.
  
To bridge this gap, we propose the News Extracted Topic-Content and Commonsense Rationale Model (\factguard), a comprehensive framework for robust fake news detection. \factguard leverages the semantic understanding capabilities of LLMs to extract event-centric information, thereby reducing the influence of textual style. By integrating LLM-generated commonsense reasoning with advanced content extraction and dynamic reliability assessment, \factguard enables a more accurate assessment of news veracity. Furthermore, the framework incorporates knowledge distillation, thereby supporting practical deployment across resource-rich~\footnote{Resource-rich refers to settings where LLMs are available and can be employed without restrictions, enabling improved fake news detection performance.}, cold-start, and resource-constrained settings and achieving a balance between accuracy and efficiency.

Specifically, in journalism communication theory, news come from the real-world events~\cite{galtung1965structure} and style rewriting is a common deceptive strategy in fake news~\cite{potthast2018stylometric} to accelerate news dissemination. So \factguard first utilizes LLMs with carefully designed prompts to extract the core topic and principal content from news articles. This process filters out stylistic noise and preserves essential event information. To ensure the quality and relevance of the extracted content, we introduce a two-stage constraint mechanism: a text similarity metric is applied during extraction to maintain consistency with the original news, followed by an information density metric that evaluates informativeness post-extraction. The resulting event-centric content, being more objective and concise, is then semantically compared with LLM-generated commonsense rationale to enhance fake news detection.
To further improve detection reliability, \factguard incorporates an LLM Rationale usability module that treats the LLM as an advisor and dynamically assesses the trustworthiness of its advice via a dual-branch structure. One branch adaptively controls the influence of LLM-based judgments, while the other emphasizes potential conflicts or ambiguities identified through commonsense reasoning, enabling the model to better address complex cases.
In addition, to support deployment in cost- and efficiency-sensitive scenarios, we introduce a knowledge distillation scheme~\cite{hinton2015distilling}. This mechanism transfers knowledge from the full \factguard model to a lightweight variant, \factguard-D, which delivers efficient inference while preserving strong detection performance.

Extensive experiments on two widely used real-world fake news detection datasets, GossipCop~\cite{shu2020fakenewsnet} and Weibo21~\cite{nan2021mdfend}, demonstrate that \factguard consistently outperforms state-of-the-art baselines across multiple key metrics, including accuracy and robustness. The distilled variant, \factguard-D, also achieves competitive results with minimal performance degradation, confirming its practicality and robustness in resource-constrained settings.
Our contributions are as follows:
\begin{itemize}    
    \item We propose \factguard, a novel framework that effectively mitigates the impact of textual style on fake news detection and enables robust integration of LLM-based reasoning. To accommodate diverse practical needs, \factguard is suitable for resource-rich scenarios, whereas its distilled variant, \factguard-D, is tailored for cold-start and resource-constrained scenarios.
    
    \item We develop an LLM-based news extraction approach that leverages semantic understanding to obtain key topics and event content, thereby reducing style interference and enabling alignment with commonsense reasoning.
    
    \item We introduce an LLM rationale usability module, which dynamically adjusts the influence of LLM advice through a dual-branch structure based on their reliability and the presence of commonsense conflicts, ensuring the effective and adaptive use of LLM knowledge.

    \item We conduct extensive experiments on the GossipCop and Weibo21 datasets, which demonstrate the effectiveness and efficiency of the proposed \factguard and \factguard-D models.
\end{itemize}

\section{Related Work}

\subsection{Traditional Fake News Detection}

Fake news detection focuses on the early identification of misinformation, primarily based on the textual content available at publication~\cite{qian2018neural}. Early methods mainly relied on machine learning models with handcrafted features, including keywords, grammatical errors~\cite{granik2017fake}, shallow linguistic patterns~\cite{wang2017liar}, and statistical cues such as text length, capitalization, and punctuation~\cite{castillo2011information}. With advances in deep learning, LSTM-based approaches were introduced to capture linguistic differences in news with satirical or rumor styles~\cite{rashkin2017truth}, and some studies explored sentiment-related features~\cite{ajao2019sentiment,giachanou2019leveraging}. However, these methods fundamentally rely on surface-level features, making them vulnerable to variations in writing style.

To address these limitations, style modeling with pretrained language models such as BERT and RoBERTa~\cite{przybyla2020capturing} has become common in fake news detection. Nevertheless, approaches relying solely on textual features remain inadequate for combating increasingly sophisticated misinformation. Recent studies have therefore incorporated background knowledge beyond textual content, including social context~\cite{shu2019defend,cui2022meta}, social emotion~\cite{zhang2021mining}, news environment~\cite{sheng2022zoom}, and external knowledge~\cite{hu2022chef}. While small language models (SLMs) offer certain improvements, their limited knowledge and capacity continue to constrain further progress in fake news detection.

\begin{figure*}[t]
\centering
\includegraphics[width=1.0\textwidth]{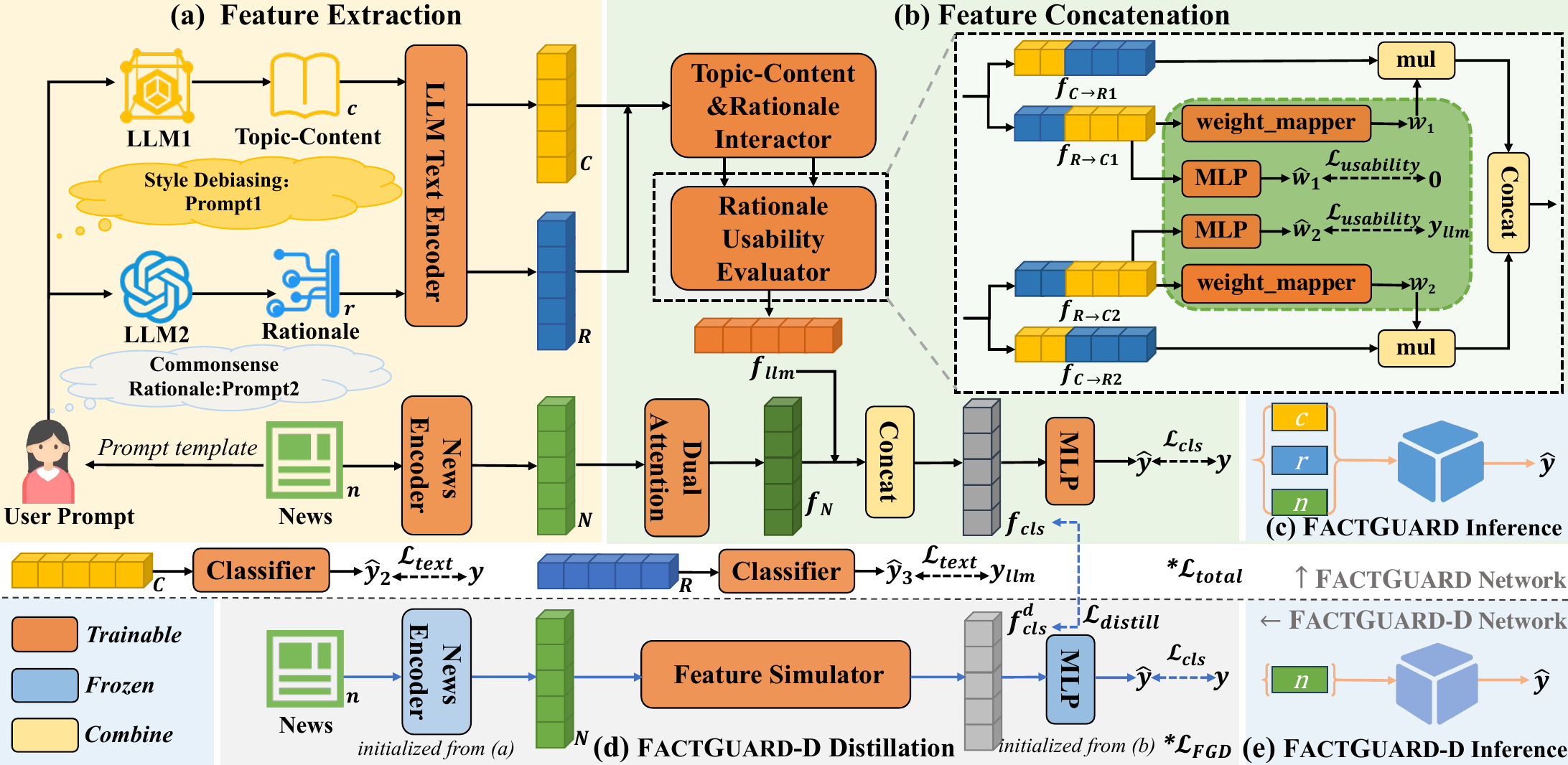}
\setlength{\abovecaptionskip}{0pt}
\setlength{\belowcaptionskip}{-5pt}
\caption{
Overview of \factguard and \factguard -D. \factguard main consists of two modules: \textbf{(1) Feature Extraction}, which identifies topic content and enables commonsense reasoning for each news article using an LLM. The resulting features and the original text are encoded for downstream processing. \textbf{(2) Feature Concatenation}, which adaptively integrates LLM-derived features with news content via a cross-attention mechanism and the Rationale Usability Evaluator, followed by MLP-based classification. After training, knowledge distillation yields a lightweight \factguard-D without LLMs' advice.
}

\label{Model}
\end{figure*}

\subsection{LLM-based Fake News Detection}
Recent studies have leveraged the strong language generation and comprehension capabilities of LLMs for content generation and enhancement in fake news detection. For content generation, LLM-Fake analyzes the psychological motivations behind LLM-generated fake news and constructs the MegaFake dataset~\cite{wang2024megafake}; SheepDog employs LLM-generated multi-style samples as adversarial data to improve detector robustness~\cite{wu2024fake}. For content enhancement, ARG leverages prompt engineering to guide LLM in multi-perspective analysis, with SLM integrating the final judgment~\cite{hu2024bad}. LEKD incorporates offline LLM knowledge via semantic graph alignment and knowledge distillation~\cite{chen2025enhancing}. TED introduces a structured multi-agent debate mechanism, enabling LLM to reason from diverse perspectives~\cite{liu2025truth}. GenFEND simulate readers with different identities to generate diverse comments, thereby providing additional information for early-stage detection~\cite{nan2024let}.

Despite these advances, the fundamental problem of style sensitivity remains unresolved. While LLM-based content generation increases sample diversity, it often fails to capture event semantics and does not fully eliminate style-related interference~\cite{wang2024megafake,wu2024fake}. In addition, augmented data generated by LLMs is often not effectively integrated with the detection backbone, resulting in limited overall improvement~\cite{hu2024bad}. Excessive reliance on LLMs and their high inference costs further constrain practical deployment~\cite{hu2024bad,liu2025truth, nan2024let}, particularly in cold-start or resource-constrained scenarios.

In summary, vulnerability to writing style, insufficient event-centric modeling, and the challenge of efficiently integrating LLM capabilities remain open problems. Motivated by these gaps, this paper proposes a new framework, \factguard, that systematically addresses style interference, leverages LLMs' commonsense reasoning capacity, and supports efficient deployment for fake news detection.

\section{Framework}

\subsection{Preliminary}
\subsubsection{Problem Formulation.}
In resource-rich scenarios, we consider a dataset $\mathcal{D}_{\text{news}} = \{(n_i, c_i, r_i)\}_{i=1}^N$, where $n_i$ denotes the news text, while $c_i$ and $r_i$ represent the event-based topic-content and the commonsense rationale extracted by an LLM. The goal for each news item $D_i = (n_i, c_i, r_i)$ is to predict $\hat{y}_i \in \{0, 1\}$, where 1 indicates fake news and 0 indicates true news, by integrating the original text with LLM-generated features. The detection model (i.e., \factguard) encodes $n_i$ via a dual-attention module $f_\theta(n_i)$, fuses $c_i$ and $r_i$ with cross-attention and usability assessment $g_\phi(\mathrm{CA}(c_i, r_i))$, and concatenates both for final prediction:
\begin{equation}
    \hat{y}_i = \mathrm{MLP}\left([f_\theta(n_i);\, g_\phi(\mathrm{CA}(c_i, r_i))]\right)
\end{equation}
where [; ] denotes vector concatenation.
To further address cold-start and resource-constrained scenarios, we distill an efficiency and resource-friendly student model (i.e., \factguard-D) from the teacher model \factguard, where the \factguard-D only takes $n_i$ as input and predicts its label $\hat{y}_i$:
\begin{equation}
\hat{y}_i = f_{\psi}(n_i)    
\end{equation}

\subsubsection{Notations.}
All notations used throughout this paper are summarized in Table~\ref{Table 3-1} in Appendix~\ref{sec:notation}.

\subsection{\factguard Overview}
Figure~\ref{Model}(a–c) illustrates the main modules as well as the training and inference procedures of \factguard. The core goal of \factguard is to achieve style debiasing and fully exploit the capabilities of LLMs for robust fake news detection. For each news item $n$, the model first employs the LLM to extract topic-content information $c$ as well as commonsense rationale $r$. These elements, together with the original news text, are encoded using an SLM (see Figure~\ref{Model}(a)). The Topic-Content\&Rationale Interactor enables deep feature interaction between the extracted topic-content and the commonsense rationale, while the Rationale Usability Evaluator adaptively assigns weights to the LLM-provided advice. The resultant interacted features $f_{llm}$ are then aggregated with the news features $f_N$. The fused representations are utilized for veracity prediction for $n$ (see Figure~\ref{Model}(b)). During training, three loss functions—$L_{cls}$, $L_{usability}$, and $L_{text}$—are employed to optimize model parameters. Once \factguard is well trained, it can be used to predict the veracity of the unseen news sample $n$ by leveraging the inputs $c$, $r$, and $n$ (see Figure~\ref{Model}(c)).

\subsection{Feature Extraction}
Pretrained SLMs such as BERT or RoBERTa are trained on large-scale datasets in an unsupervised fashion, enabling them to generate high-dimensional contextual representations well-suited for various downstream tasks. To effectively extract information features, we employ these models as text encoders within our framework. Specifically, for a given news item $n$, the extracted topic-content $c$, and the commonsense rationale $r$, we denoted them as $N$ (news), $C$ (topic-content), and $R$ (commonsense rationale), respectively.

\subsection{Feature Concatenation}
This module aims to obtain high-quality representations for both the LLM-generated augmented information and the original news content, facilitating their effective integration and collaboration as the foundation for fake news detection. Feature integration is performed via concatenation of the respective representations, enabling subsequent modules to leverage comprehensive contextual information.

\subsubsection{Topic-Content\&Rationale Interactor.}

To enable comprehensive feature exchange between the LLM-extracted topic-content and commonsense rationale, we introduce a dual cross-attention module based on multi-head attention. The computation is formulated as follows:
\begin{equation}
\text{head}_i = \text{softmax} \left( \frac{Q_i K_i^\top}{\sqrt{d_k}} \right) V_i,
\end{equation}
\begin{equation}
\text{CA}(Q, K, V) = \text{Concat}(\text{head}_1, \ldots, \text{head}_h) W^O,
\end{equation}
where $Q_i = Q W_i^Q$, $K_i = K W_i^K$, and $V_i = V W_i^V$. Here, $d_k$ is the dimension of each attention head, $h$ is the number of heads, and $W^O$ is the output projection matrix. Given topic-content $c$ and commonsense rationale $r$, after embedding, the interactions are computed as:
\begin{equation}
f_{C \rightarrow R} = \text{AvgPool}\left(\text{CA}(C, R, R)\right),
\end{equation}
\begin{equation}
f_{R \rightarrow C} = \text{AvgPool}\left(\text{CA}(R, C, C)\right),
\end{equation}
where $\text{AvgPool}(\cdot)$ denotes average pooling applied over the token representations output by the cross-attention layer. $f_{C \rightarrow R}$ represents the LLM advice feature vector, and $f_{R \rightarrow C}$ serves as a weighting factor in the Rationale Usability Evaluator.

\subsubsection{Rationale Usability Evaluator.}
Directly enforcing consistency between LLM judgments and ground-truth labels may result in the loss of valuable complementary features provided by the LLMs. To address this, we propose a rationale usability evaluation module to dynamically adjust the fusion weights of LLM features, thereby maximizing the utility of LLM-generated knowledge. This module adopts a dual-branch MLP structure: one branch reduces its contribution as the LLMs' direct detection capability is limited, while the other increases its contribution as commonsense reasoning can provide more effective information when it identifies contradictions or uncertainty. A three-layer MLP (weight\_mapper) maps the feature vectors $f_{R \rightarrow C_i}$ to fusion weights $w_i$ as follows:
\begin{equation}
w_i = \mathrm{sigmoid({weight\_mapper}}(f_{R \rightarrow C_i})),\quad i=1,2.
\end{equation}

The final LLM feature representation $f_{llm}$ is then computed as:
\begin{equation}
f_{llm} = [w_1 \cdot f_{C \rightarrow R1}; w_2 \cdot f_{C \rightarrow R2}],
\end{equation}
where $w_1$ and $w_2$ are the fusion weights for the two branches, and $f_{C \rightarrow R1}$, $f_{C \rightarrow R2}$ denote their respective interaction features.

\subsubsection{Dual Attention Fusion.}

To further improve the expressiveness and robustness of features derived from BERT or RoBERTa, we introduce a linear attention mechanism that adaptively assigns higher weights to salient tokens, thereby suppressing irrelevant or noisy information:
\begin{equation}
\mathrm{Attn}(X) = \sum_{t=1}^{T} \mathrm{softmax}(W x_t + b) \cdot x_t,
\end{equation}
where $x_t$ denotes the input feature at position $t$, $W$ and $b$ are learnable parameters, and $T$ is the token sequence length.  
To enhance model robustness, a dual-branch architecture is adopted, where the same linear attention module is applied in parallel to both branches. The outputs of the two branches are then averaged to obtain the final news feature representation:
\begin{equation}
f_N = \frac{\mathrm{Attn}(N) + \mathrm{Attn}(N)}{2}.
\end{equation}

\subsubsection{Feature Concatenation.}
Based on the outputs obtained in the previous step, the news feature vector $f_N$ and the LLM-enhanced feature vector $f_{llm}$ are summed to facilitate the final prediction. For each news item $n$ with label $y \in \{0,1\}$, these vectors are combined to produce the final feature representation $f_{cls}$, computed as:
\begin{equation}
f_{cls} = [f_N; f_{llm}].
\end{equation}

$f_{cls}$ is subsequently input into an MLP classifier to predict the veracity label:
\begin{equation}
\hat{y} = \mathrm{MLP}(f_{cls}).
\end{equation}

\subsection{Training}

\subsubsection{Data Process.}
To enhance semantic understanding and reduce the influence of writing style, LLMs is leveraged to extract the topic-content of each news article. Additionally, commonsense rationale analysis is performed by the LLMs to identify and judge content that may contradict commonsense. Due to page limitations, detailed information on the data processing procedure is provided in Appendix~\ref{sec:process}.

\subsubsection{Objective.}
The \factguard method is designed with three principal objectives: \ding{172} to achieve accurate prediction of news veracity; \ding{173} to effectively integrate model recommendations and fully leverage the capabilities of LLMs; and \ding{174} to enhance the representation of information augmentation provided by LLMs. Accordingly, the overall objective loss function is defined as a weighted sum of the prediction loss, the LLM rationale usability loss, and the information augmentation representation loss.

To improve the final detection performance, the Binary Cross-Entropy (BCE) classification loss is computed to guide the model in accurately identifying fake news:
\begin{equation}
\mathcal{L}_{cls} = \text{BCE}(\hat{y}, y).
\end{equation}

To supervise the learning of LLM features, the supervision signals for the weights are set as 0 for one branch and $y_{llm}$ for the other:
\begin{equation}
\hat{w}_i=\mathrm{sigmoid}(\mathrm{MLP}(f_{C \rightarrow R_i})), \quad i=1,2,
\end{equation}
\begin{equation}
\mathcal{L}_{usability} = \text{BCE}(\hat{w}_{1}, 0)+\text{BCE}(\hat{w}_{2}, y_{llm}).
\end{equation}

To enhance the utility of LLM-generated augmentations, we employ an auxiliary task that aligns extracted semantics and commonsense reasoning with ground-truth labels and LLM veracity judgments. Augmented representations are fed into a classifier composed of a linear attention and an MLP head (without sigmoid), optimized by cross-entropy (CE) loss:
\begin{equation}
\hat{y}_2=\mathrm{Classifier(}C),\quad \hat{y}_3=\mathrm{Classifier(}R),
\end{equation}
\begin{equation}
\mathcal{L}_{text}= \text{CE}(\hat{y}_2, y) + \text{CE}(\hat{y}_3, y_{llm}).
\end{equation}

The total loss function is a weighted sum of the aforementioned terms:
\begin{equation}
\mathcal{L}_{total}
= \mathcal{L}_{cls} + \alpha \frac{\mathcal{L}_{usability}}{2} + \beta \frac{\mathcal{L}_{text}}{2},
\end{equation}
where $\alpha$ and $\beta$ are hyperparameter weights, and the division by 2 is used to average the two sub-losses, ensuring balanced contributions from each component in the overall loss.

\subsection{Inference}
In resource-rich scenarios, LLMs are leveraged via prompt engineering to extract the topic-content, and commonsense rationales of news articles. The extracted outputs, together with the original news text, are subsequently fed into the well-trained frozen \factguard model for veracity prediction.

 Due to page limitations, further training and inference details for \factguard are provided in Algorithm~\ref{alg:alg1} and Algorithm~\ref{alg:alg3} in Appendix~\ref{sec:algorithm}.

\subsection{\factguard-D}
\subsubsection{Distillation.}
Directly invoking LLMs for each prediction in \factguard is impractical in resource-constrained or latency-sensitive cold-start scenarios due to the substantial overhead of real-time LLM prompting for text extraction and commonsense reasoning. To address this, we develop a llm-free student model via knowledge distillation from the trained \factguard, following a teacher–student paradigm~\cite{hu2024bad}. The core idea is to transfer and internalize the teacher model’s reasoning knowledge into a parameterized lightweight student network. Specifically, as illustrated in Figure~\ref{Model}(d), the student model's news encoder and classifier are initialized from the trained \factguard. To acquire the teacher's reasoning capabilities, a feature simulator is implemented as a four-layer Transformer encoder and a linear attention module to internalize the teacher’s knowledge. In addition to the standard prediction loss $\mathcal{L}_{cls}$ as in \factguard, the student model is further supervised by a feature distillation loss $\mathcal{L}_{distill}$, which encourages the student’s feature representation $f^{d}_{cls}$ to approximate that of the teacher $f_{cls}$ by minimizing the mean squared error (MSE) between them:
\begin{equation}
\mathcal{L}_{distill} = \text{MSE}(f^{d}_{cls}, f_{cls}).
\end{equation}

\subsubsection{Inference.}
In cold-start and resource-constrained scenarios, the \factguard-D model operates exclusively on the original news text $n$, achieving fast predictions with only a slight reduction in accuracy. 

Due to page limitations, further distillation training and inference details for \factguard-D are provided in Algorithm~\ref{alg:alg2} and Algorithm~\ref{alg:alg4} in Appendix~\ref{sec:algorithm}.

\begin{table*}[ht]
\small
\centering
{
\begin{tabular}{llcccccccc}
\toprule 
\multirow{2}{*}{\textbf{Group}} & \multirow{2}{*}{\textbf{Model}} & \multicolumn{4}{c}{\textbf{Weibo21}} & \multicolumn{4}{c}{\textbf{GossipCop}} \\
\cmidrule(lr){3-6} \cmidrule(lr){7-10} 
& & $\text{macF1}$ & $\text{Acc.}$ & $\text{F1}_\text{real}$ & $\text{F1}_\text{fake}$ & $\text{macF1}$ & $\text{Acc.}$ & $\text{F1}_\text{real}$ & $\text{F1}_\text{fake}$ \\
\midrule

\multirow{4}{*}{G1} 
& GPT-3.5-turbo* & 0.725 & 0.734 & 0.774 & 0.676 & 0.702 & 0.813 & 0.884 & 0.519 \\ 
& GPT-4o-mini\# & 0.725 & 0.746 & 0.780 & 0.670 & 0.691 & 0.845 & 0.909 & 0.472 \\ 
& ChatEval-o\# & 0.694 & 0.717 & 0.778 & 0.611 & 0.733 & 0.860 & 0.919 & 0.546 \\ 
& ChatEval-s\#& 0.694 & 0.719 & 0.780 & 0.608  & 0.738 & 0.869 & 0.923 & 0.553\\

\midrule
\multirow{4}{*}{G2} 
& BERT* & 0.753 & 0.754 & 0.769 & 0.737 & 0.765 & 0.862 & 0.916 & 0.615 \\ 
& RoBERTa & 0.753 & 0.755 & 0.775 & 0.731 & 0.765 & 0.862 & 0.916 & 0.613 \\ 
& EANN* & 0.754 & 0.756 & 0.773 & 0.736 & 0.763 & 0.864 & 0.918 & 0.608 \\ 
& Publisher-Emo* & 0.761 & 0.763 & 0.784 & 0.738 & 0.766 & 0.868 & 0.920 & 0.611 \\ 
& ENDEF* & 0.765 & 0.766 & 0.779 & 0.751 & 0.768 & 0.865 & 0.918 & 0.618 \\

\midrule
\multirow{5}{*}{G3} 
& Bert + Rationale* & 0.767 & 0.769 & 0.787 & 0.748 & 0.777 & 0.870 & 0.921 & 0.633 \\ 
& SuperICL* & 0.757 & 0.759 & 0.779 & 0.734 & 0.736 & 0.864 & 0.920 & 0.551 \\ 
&Bert + GenFEND  &0.755 & 0.760 & 0.791& 0.719& 0.764 & 0.875 & 0.926 & 0.603 \\
&Roberta + GenFEND & 0.771 & 0.774 & 0.796 & 0.747 &  0.770 & 0.866 & 0.919 & 0.621 \\
& ARG*& 0.784 & 0.786 & 0.804 & 0.764 & 0.790 & 0.878 & 0.926 & 0.653 \\ 
& TED\# & \underline{0.795} & \underline{0.798} & \underline{0.815} & \underline{0.774} & \underline{0.803} & \textbf{0.892} & 0.932 & \underline{0.674} \\ 
& \textbf{Ours} & \textbf{0.801} & \textbf{0.804} & \textbf{0.824} & \textbf{0.777} & \textbf{0.805} & \textbf{0.892} & \textbf{0.935} & \textbf{0.675} \\ 
\midrule \midrule
\multirow{2}{*}{G4} 
& ARG-D* & 0.771 & 0.772 & 0.785 & 0.756 & 0.778 & 0.870 & 0.921 & 0.634 \\ 
& \textbf{Ours} & 0.788 & 0.790 & 0.807 & 0.769 & 0.790 & 0.888 & \underline{0.933} & 0.647 \\

\bottomrule
\end{tabular}
\caption{Performance comparison on Weibo21 and GossipCop datasets across four metrics, i.e.,$\text{macF1}$, $\text{Accuracy}$, $\text{F1}_\text{real}$, and $\text{F1}_\text{fake}$. The highest result in each category is \textbf{bolded} and the second highest result is \underline{underlined}. In the results table, * means the result is from ~\cite{hu2024bad} and \# means the result is from~\cite{liu2025truth}.}
\label{Table 4-2}}
\end{table*}

\section{Experiments}
This section presents comprehensive experimental studies of the proposed \factguard and \factguard-D models. We first introduce the experimental setup (for detailed settings, please refer to Appendix~\ref{app_sec:experimental_setup}). Subsequently, we compare \factguard with a wide range of baselines, conduct ablation studies to assess the contribution of each model component, analyze parameter sensitivity, and discuss challenges associated with LLM-based text extraction.

\subsection{Setup}
\subsubsection{Datasets.}
We employ the Weibo21 (Chinese)~\citep{nan2021mdfend} and GossipCop (English)~\citep{shu2020fakenewsnet} for evaluation. Both datasets are preprocessed by deduplication and temporal splitting, following established practices~\citep{zhu2022generalizing,mu2023s,hu2024bad}, to mitigate the risk of data leakage and prevent overestimation of SLM performance. In addition, we also utilize the commonsense rationales from \cite{hu2024bad}.

\subsubsection{Baselines.}
Recent fake news detection methods predominantly rely on LLMs and SLMs, and can be categorized into four groups. Among them, we involved 14 representative baselines in this work. The first group (G1) comprises LLM-only methods, including GPT-3.5-turbo~\cite{openai2023gpt35}, GPT-4o-mini~\cite{openai_gpt4o_mini_blog}, ChatEval-o (one-by-one strategy)~\cite{chan2024chateval}, and ChatEval-s (Simultaneous-Talk strategy)~\cite{chan2024chateval}. The second group (G2) consists of SLM-only methods, such as BERT~\cite{devlin2019bert}, RoBERTa~\cite{liu2019roberta}, EANN~\cite{wang2018eann}, Publisher-Emo~\cite{zhang2021mining}, and ENDEF~\cite{zhu2022generalizing}. The third group (G3) includes LLM-SLM methods, such as BERT + Rationale~\cite{hu2024bad}, SuperICL~\cite{zhong2023can}, ARG~\cite{hu2024bad}, BERT + GenFEND\cite{nan2024let}, RoBERTa + GenFEND and TED~\cite{liu2025truth}. The fourth group (G4) comprises methods employing model distillation, such as ARG-D~\cite{hu2024bad}. 

\subsubsection{Metrics.}
We evaluate performance using four metrics: Accuracy ($\mathrm{Acc.}$), $\mathrm{F1}_\mathrm{real}$, $\mathrm{F1}_\mathrm{fake}$, and Macro-F1 ($\mathrm{macF1}$). 

\subsubsection{Implementation Details.}
We utilize bert-base-chinese\footnote{\url{https://huggingface.co/google-bert/bert-base-chinese}}~\cite{devlin2019bert} as the text encoder for Chinese \factguard model and roberta-base\footnote{\url{https://huggingface.co/FacebookAI/roberta-base}}~\cite{liu2019roberta} for English \factguard model. For the Weibo21 and GossipCop dataset, topic-content extraction is performed using locally deployed DeepSeek-R1-Distill-Llama-8B
\footnote{\url{https://huggingface.co/deepseek-ai/DeepSeek-R1-Distill-Llama-8B}}~\cite{deepseekai2025} and SOLAR-10.7B-Instruct-v1.0-uncensored\footnote{\url{https://huggingface.co/upstage/SOLAR-10.7B-Instruct-v1.0}}~\cite{kim2024solar}~\cite{kim2024solar}, respectively. Commonsense reasoning modules for both datasets are adopted from prior work~\cite{hu2024bad}. We employ the AdamW optimizer with an initial learning rate of $2e-4$ and a weight decay of $5e-5$. Early stopping with a patience of 5 epochs is applied to prevent overfitting. All experiments are conducted on a single NVIDIA A100 (40GB) GPU with a fixed random seed of 3759, PyTorch version 1.13.0.

\subsection{Comparative Results}
To evaluate the effectiveness of the proposed \factguard model, we conduct systematic experiments on the Weibo21 and GossipCop datasets. As shown in Table~\ref{Table 4-2}, \factguard consistently outperforms all baseline methods across multiple evaluation metrics, achieving the best results across both datasets. These results underscore the superior cross-lingual generalization and stability of the proposed model.

\begin{table*}[t]
\centering
\small
{
\begin{tabular}{llcccccccc}
\toprule
 \multirow{2}{*}{\textbf{Group}} &  \multirow{2}{*}{\textbf{Model}} & \multicolumn{4}{c}{\textbf{Weibo21}} & \multicolumn{4}{c}{\textbf{GossipCop}} \\
\cmidrule(lr){3-6} \cmidrule(lr){7-10}
 & & macF1 & Acc. & F1$_\text{real}$ & F1$_\text{fake}$ & macF1 & Acc. & F1$_\text{real}$ & F1$_\text{fake}$ \\
\midrule

\multirow{1}{*}{G1} 
    & \textbf{\factguard} & \textbf{0.801} & \textbf{0.804} & \textbf{0.824} & \textbf{0.777} & \textbf{0.805} & \textbf{0.892} & \textbf{0.935} & \textbf{0.675} \\

\midrule
\multirow{3}{*}{G2} 
    & News & 0.768 & 0.769 & 0.784 & 0.751 & 0.765 & 0.862 & 0.916 & 0.613 \\
    & Topic-Content & 0.690 & 0.691 & 0.708 & 0.672 & 0.769 & 0.861 & 0.915 & 0.624 \\
    & Commonsense & 0.678 & 0.685 & 0.728 & 0.627 & 0.698 & 0.832 & 0.899 & 0.498 \\

\midrule
\multirow{3}{*}{G3} 
    & w/o News & 0.718 & 0.722 & 0.753 & 0.683 & 0.773 & 0.873 & 0.924 & 0.623 \\
    & w/o Topic-Content & 0.772 & 0.774 & 0.794 & 0.750 & 0.779 & 0.875 & 0.925 & 0.632 \\
    & w/o Commonsense & 0.770 & 0.773 & 0.797 & 0.743 & 0.779 & 0.871 & 0.922 & 0.633 \\
    
\midrule
\multirow{2}{*}{G4} 
    & w/o llm-usability& 0.778 & 0.780 & 0.794 & 0.763 & 0.780 & 0.878 & 0.927 & 0.633 \\
    & use ARG-usefulness  & 0.782 & 0.782 & 0.793 & 0.770 & 0.775 & 0.872 & 0.923 & 0.628 \\
    
\bottomrule
\end{tabular}
}

\caption{ An ablation study of \factguard on the Weibo21 and GossipCop datasets evaluates the contribution of each model component. Specifically, G2 assesses the predictive power of individual input features (original news, LLM-extracted topic-content, LLM commonsense rationale judgment); G3 quantifies the impact of ablating each core input module; and G4 investigates the effects of removing LLM rationale usability module on overall performance and cross-lingual generalization.}

\label{Table 4-3}
\end{table*}

\subsubsection{Weibo21.} On the Weibo21 dataset, \factguard outperforms the strongest baseline TED by 0.8\% in accuracy and 0.9\% in $\text{F1}_\text{real}$, along with notable macF1 gains. These improvements arise from its LLM-driven topic–content extraction, dual-branch reasoning, and efficient rationale usability module, which jointly combine SLM efficiency with LLM reasoning capacity. Compared with TED’s complex multi-agent debate framework, \factguard achieves higher accuracy with only two simple prompts, significantly reducing computational resource and inference costs. The distilled variant, \factguard-D, further surpasses ARG and ARG-D, demonstrating effective compression and strong practical performance.

\subsubsection{GossipCop.} On the GossipCop dataset, although the improvements are smaller, \factguard still attains the best macF1 (0.805) and $\text{F1}_\text{real}$ (0.935). Under limited computational resources, \factguard-D also achieves lower misclassification rates, confirming its broad applicability. The performance differences across datasets mainly result from class imbalance (more fake news in Weibo21, more real news in GossipCop), yet \factguard remains stable and generalizable across languages.

In summary, both \factguard and \factguard-D demonstrate strong performance, cross-lingual robustness, and model compression capability, making them efficient and reliable solutions for multilingual fake news detection.

\subsection{Ablation Study}
We conduct ablation studies on the GossipCop and Weibo21 datasets to evaluate the effectiveness of each module in the proposed \factguard framework, focusing on the LLM-based topic-content extraction, commonsense rationale judgment, and usability evaluation modules. The ablation experiments are organized into three groups:

As shown in Table~\ref{Table 4-3}, the main findings are as follows: (1) Removing the original news representation yields the largest reduction in overall performance, confirming its indispensable role as the foundation for fake news detection. (2) Removing the LLM-based topic-content extraction module results in a notable drop in macF1 and $\text{F1}_\text{real}$, highlighting its importance in capturing essential event information and reducing stylistic noise. (3) Excluding the LLM commonsense rationale module further degrades performance, demonstrating its value in improving factual consistency and reasoning. (4) Omitting the usability evaluation module or use ARG's LLM usefulness module also leads to decreased performance, underscoring its role in aligning LLM inference with robust news representations. (5) The LLM-extracted topic-content module and the LLM commonsense rationale module need to be used in conjunction to achieve the maximum performance improvement. Overall, these results validate that each component is essential for \factguard’s strong performance in multilingual fake news detection.

In addition to the main ablation experiments, we performed both grid search and random search to determine optimal loss weight parameters for the multi-objective loss functions in the \factguard model. For the Weibo21 configuration, the best results were achieved with $\alpha = 0.40$ and $\beta = 0.16$, while for the GossipCop configuration, the optimal values were $\alpha = 0.50$ and $\beta = 0.58$. To facilitate effective distillation, we set the distillation coefficient $\lambda$ in the loss function to 8 for both the Chinese and English models in \factguard-D. Additional experimental details are provided in Appendix~\ref{sec:sensitivity_study}.

\section{Conclusion}
We propose \factguard, a model that leverages LLMs for semantic understanding and commonsense reasoning to improve fake news detection performance. By extracting topic and core content and employing a usability evaluation module in commonsense rationale, \factguard effectively reduces style bias and integrates LLM-generated judgments. For cold-start and resource-limited scenarios, the distilled variant \factguard-D is optimized for efficiency and resources. Experiment results on Weibo21 and GossipCop datasets show that \factguard outperforms baselines with each module proven effective by ablation studies, while \factguard-D achieves a strong balance between accuracy and speed.

\noindent\textbf{Future Work.}
Future directions include: (1) developing customized methods for Chinese and English fake news; (2) optimizing the model for edge deployment; (3) enhancing interpretability of usability evaluation module to improve transparency and credibility; (4) exploring the role of text style at different stages of news dissemination detection; and (5) considering the benchmark data contamination of the employed LLMs, and extending cross-domain adaptation across emerging platforms and multimodal signals.

\section{Acknowledgments}
This work was supported in part by Scientific Research and Innovation Project of Postgraduate Students in the Academic Degree of Yunnan University under KC-252512080, in part by the National Natural Science Foundation of China under Grants 62162067 and 82360280, in part by the Yunnan Province Special Project under Grant 202403AP140021 and in part by the Yunnan Fundamental Research Project under Grant 202401AT070474.

\bibliography{aaai2026}

\newpage
\appendix

\section{Appendix Overview}
This appendix provides supplementary material that could not be included in the main paper due to space constraints. Specifically, it includes:

\begin{itemize}
    \item \textbf{Section.~\ref{sec:notation}:} Introduction of all notations used throughout the paper.
    \item \textbf{Section.~\ref{sec:algorithm}:} Complete algorithmic procedures for training and inference in the proposed \factguard and \factguard-D frameworks.
    \item \textbf{Section.~\ref{sec:process}:} LLM-based news enhancement, including topic and content extraction, extraction results, and commonsense rationale-based news judgment.  
    \item \textbf{Section.~\ref{app_sec:experimental_setup}:} Details of datasets, baselines and evaluation metrics used in comparative experiments.
    \item \textbf{Section.~\ref{sec:sensitivity_study}:} Additional ablation studies on loss hyperparameters sensitivity of \factguard and \factguard -D, text encoders' choices, confidence distribution analysis and case analysis.
\end{itemize}

\section{Notation}
Notations used in this paper are summarized in Table~\ref{Table 3-1}.
\setcounter{secnumdepth}{2}  
\renewcommand\thesubsection{\Alph{section}.\arabic{subsection}} 
\label{sec:notation}
\begin{table}[ht]
\centering
\setlength{\tabcolsep}{2mm}
\begin{tabular}{ll}
\toprule
\textbf{Notation} & \textbf{Description} \\
\midrule
$c$ & LLM-extracted news topic-content \\
$r$ & LLM's commonsense rationale on news \\
$n$ & Original news content \\
$C$ & Topic-Content's token \\ 
$R$ & Commonsense rationale's token \\ 
$N$ & News' token \\ 
$f_{C \rightarrow R}$ & LLM advice features  \\
$f_{R \rightarrow C}$ & Used as the weight of $f_{C \rightarrow R}$ after score map \\
$w_i$ & Adjusted weight of LLM advice \\ 
$\hat{w_i}$ & Predicted weight of LLM advice \\ 
$f_{llm}$ & The final fused feature of $c$ and $r$ \\
$f_N$ & The final feature of $n$ \\
$f_{cls}$ & The final feature for prediction \\
$y$ & Label of news \\
$y_{llm}$ & Label of LLM's judgment \\
$\hat{y_2}$ & Prediction of $c$ \\
$\hat{y_3}$ & Prediction of $r$ \\
$L_{cls}$ & Classification Loss \\
$L_{usability}$ & LLM's advice usability loss \\
$L_{text}$ & LLM text enhanced loss\\
$L_{distill}$ & \factguard-D's distillation loss \\
$L_{total}$ & \factguard's total loss \\
$L_{FGD}$ & \factguard-D's total loss \\
$;$ & Concat corresponding dimensions of vectors \\
$*$ & Multiply vectors \\
\bottomrule
\end{tabular}
\setlength{\abovecaptionskip}{5pt}    
\caption{Notation and their corresponding descriptions.}
\label{Table 3-1}
\end{table}

\section{Algorithm}
\label{sec:algorithm}

The algorithms are divided into two parts. The first part, shown in Algorithm~\ref{alg:alg1} and Algorithm~\ref{alg:alg2}, presents the \factguard and \factguard-D training procedures. The second part, detailed in Algorithm~\ref{alg:alg3} and Algorithm~\ref{alg:alg4}, illustrates the \factguard and \factguard-D inference processes.

\begin{algorithm}[t]
\caption{\factguard Training Process}
\label{alg:alg1}
\begin{algorithmic}[1]
\Statex \textbf{Input:} News $n$, LLM extracted topic-content $c$, LLM commonsense rationale $r$, ground truth labels $y$, LLM commonsense rationale judgment $y_{llm}$, number of samples $N$, batch size $B$ 
\State \textbf{Initialization:} hyperparameters $\alpha$, $\beta$, epoch $E \gets N/B$, optimizer $\gets AdamW$
\For{$epoch = 1$ to $E$}
    \State $C \gets \text{News Encoder}(n)$
    \State $C, R \gets \text{LLM Text Encoder}(c,r)$
    \State $\hat{y}_2 \gets \text{Classifier}(C)$
    \State $\hat{y}_3 \gets \text{Classifier}(R)$
    \For{$i = 1$ to $2$}
        \State $f_{C \rightarrow Ri} \gets \text{CA}(C, R, R)$
        \State $f_{R \rightarrow Ci} \gets \text{CA}(R, C, C)$
        \State $w_i \gets \text{weight\_mapper}(f_{R \rightarrow Ci})$
        \State $\hat{w}_i \gets \text{MLP}(f_{R \rightarrow Ci})$
    \EndFor
    \State $f_{llm} \gets [w_1 * f_{C \rightarrow R1}; w_2 * f_{C \rightarrow R2}]$
    \State $f_N \gets [\text{Attn}(N) + \text{Attn}(N)] / 2$
    \State $f_{cls} \gets [f_{llm}; f_N]$
    \State $\hat{y} \gets \text{MLP}(f_{cls})$
    \State $L_{cls} \gets \text{BCE}(\hat{y}, y)$
    \State $L_{weight} \gets \text{BCE}(\hat{w}_1, 0) + \text{BCE}(\hat{w}_2, y_{llm})$
    \State $L_{text} \gets \text{CE}(\hat{y}_2, y) + \text{CE}(\hat{y}_3, y_{llm})$
    \State $L_{total} \gets L_{cls} + \alpha \frac{L_{usability}}{2} + \beta \frac{L_{text}}{2}$ 
    \State \textbf{Zero gradients:} $optimizer.zero\_grad()$
    \State \textbf{Backward pass:} $L_{total}.backward()$
    \State \textbf{Update weights:} $optimizer.step()$
\EndFor
\end{algorithmic}
\end{algorithm}

\begin{algorithm}[t]
\caption{\factguard Inference Process}
\label{alg:alg3}
\begin{algorithmic}[1]
\Statex \textbf{Input:} News $n$, LLM extracted topic-content $c$, LLM commonsense rationale $r$
\Statex \textbf{Output:} Prediction $\hat{y}$
\State \textbf{Initialization:} Frozen \factguard~model
\State $C \gets \text{News Encoder}(n)$
\State $C, R \gets \text{LLM Text Encoder}(c, r)$
\For{$i=1$ to $2$}
    \State $f_{C \rightarrow Ri} \gets \text{CA}(C, R, R)$
    \State $f_{R \rightarrow Ci} \gets \text{CA}(R, C, C)$
    \State $w_i \gets \text{MLP}(f_{R \rightarrow Ci})$
\EndFor
\State $f_{llm} \gets [w_1 \cdot f_{C \rightarrow R1};\; w_2 \cdot f_{C \rightarrow R2}]$
\State $f_N \gets [\text{Attn}(n) + \text{Attn}(n)] / 2$
\State $f_{cls} \gets [f_{llm};\; f_N]$
\State $\hat{y} \gets \text{MLP}(f_{cls})$
\State \textbf{return} $\hat{y}$
\end{algorithmic}
\end{algorithm}

\begin{algorithm}[t]
\caption{\factguard-D Training Process}
\label{alg:alg2}
\begin{algorithmic}[1]
\Statex \textbf{Input:} News $n$, LLM extracted topic-content $c$, LLM commonsense rationale $r$, ground truth labels $y$, number of samples $N$, batch size $B$ 
\State \textbf{Initialization:} Frozen \factguard model, frozen News Encoder from \factguard, frozen MLP from \factguard, hyperparameter $\lambda$, epoch $E \gets N/B$, optimizer $\gets AdamW$
\For{$epoch = 1$ to $E$}
    \State $f_{cls} \gets \text{\factguard}(c, r, n)$
    \State $C \gets \text{News Encoder}(n)$
    \State $f_{cls}^d \gets \text{Feature Simulator}(C)$
    \State $\hat{y} \gets \text{MLP}(f_{cls}^d)$
    \State $L_{cls} \gets \text{BCE}(\hat{y}, y)$
    \State $L_{distill} \gets \text{MSE}(f_{cls}, f_{cls}^d)$
    \State $L_{FGD} \gets L_{cls} + \lambda L_{distill}$
    \State \textbf{Zero gradients:} $optimizer.zero\_grad()$
    \State \textbf{Backward pass:} $L_{FGD}.backward()$
    \State \textbf{Update weights:} $optimizer.step()$
\EndFor
\end{algorithmic}
\end{algorithm}

\begin{algorithm}[ht]
\caption{\factguard-D Inference Process}
\label{alg:alg4}
\begin{algorithmic}[1]
\Statex \textbf{Input:} News $n$
\Statex \textbf{Output:} Prediction $\hat{y}$
\State \textbf{Initialization:} Frozen \factguard-D~model
\State $C \gets \text{News Encoder}(n)$
\State $f_{cls}^d \gets \text{Feature Simulator}(C)$
\State $\hat{y} \gets \text{MLP}(f_{cls}^d)$
\State \textbf{return} $\hat{y}$
\end{algorithmic}
\end{algorithm}

\begin{figure*}[t]
\centering
\includegraphics[width=1.0\textwidth]{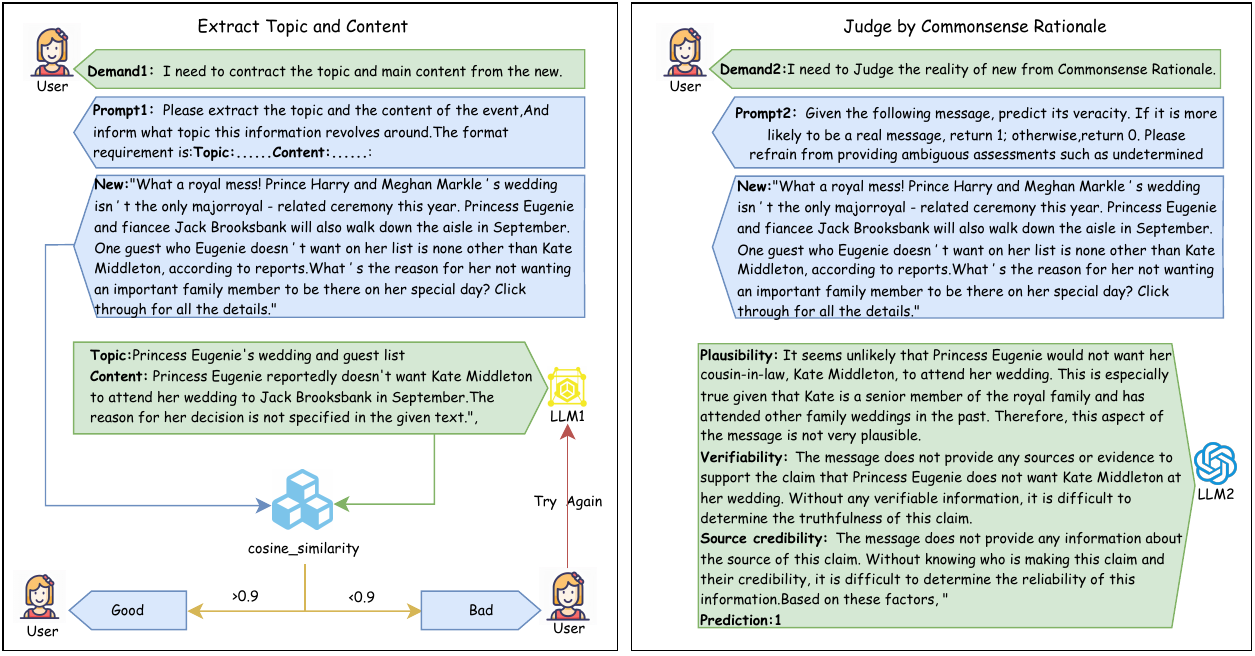}
\caption{Data processing workflow. It consists of two parts: (1) extracting topics and content to mitigate the influence of text style; and (2) performing commonsense reasoning to identify contradictions in the news and generate LLM judgments.}
\label{Prompt}
\end{figure*}
\section{Data Process}
\label{sec:process}

In this paper, we leverage an LLM for data processing through prompt engineering. The process consists of two main components: (1) extracting topics and core content to mitigate the influence of text style, and (2) generating commonsense rationales and judgments to infuse additional knowledge from the LLM. The overall workflow is illustrated in Figure~\ref{Prompt}.

\begin{figure*}[ht]  
\centering


\begin{subfigure}[t]{0.485\linewidth}
    \centering
    \includegraphics[width=\linewidth]{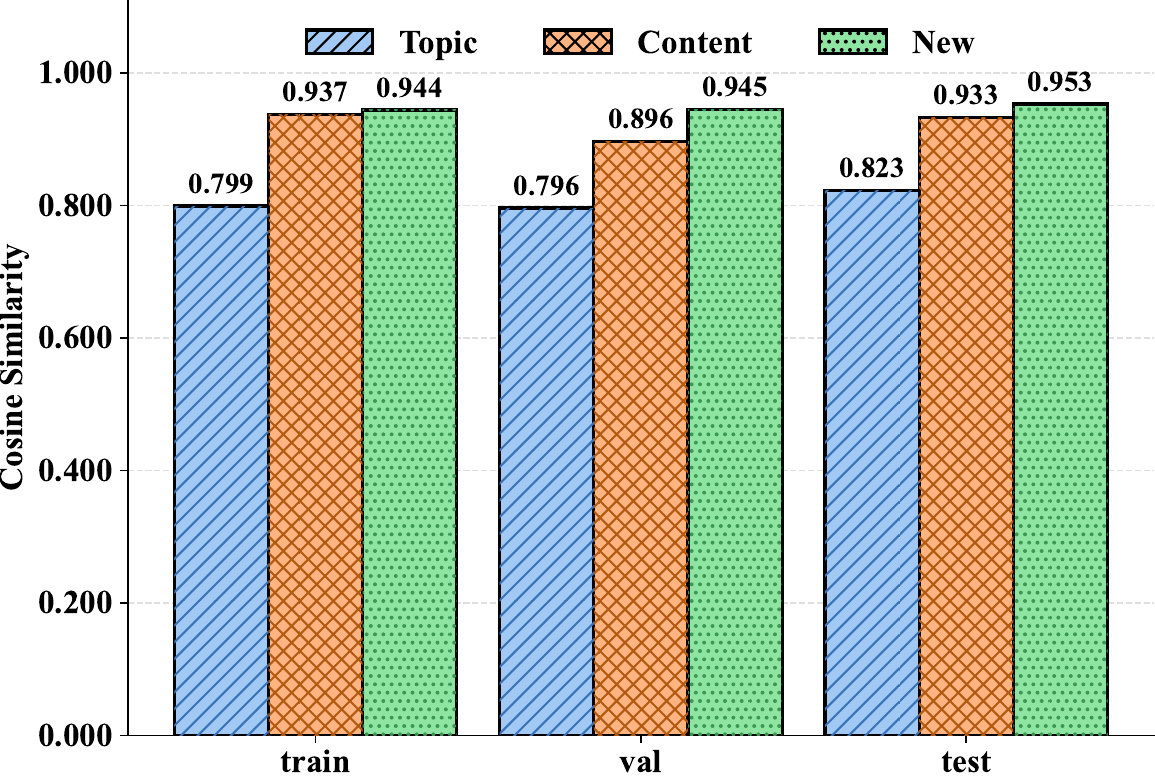}
    \caption{Weibo21 dataset similarity}
    \label{fig:Chinese_similarity}
\end{subfigure}
\hfill
\begin{subfigure}[t]{0.485\linewidth}
    \centering
    \includegraphics[width=\linewidth]{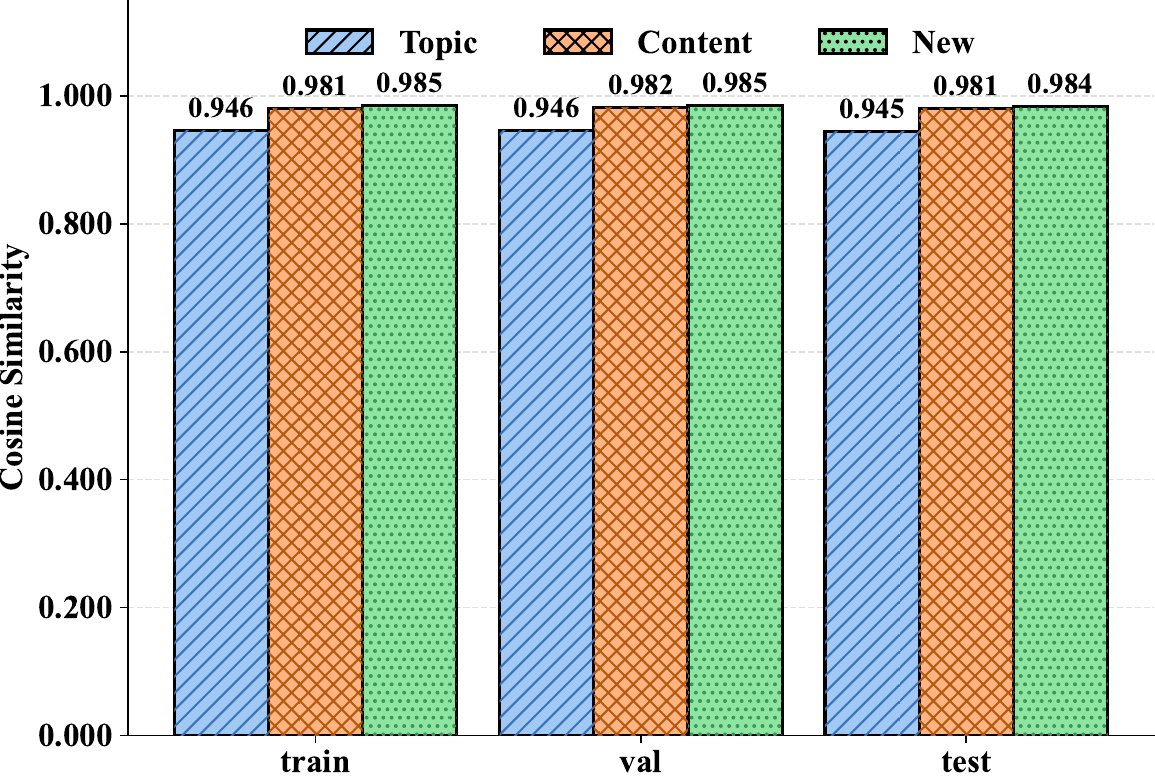}
    \caption{GossipCop dataset similarity}
    \label{fig:English_similarity}
\end{subfigure}

\begin{subfigure}[t]{0.485\linewidth}
    \centering
    \includegraphics[width=\linewidth]{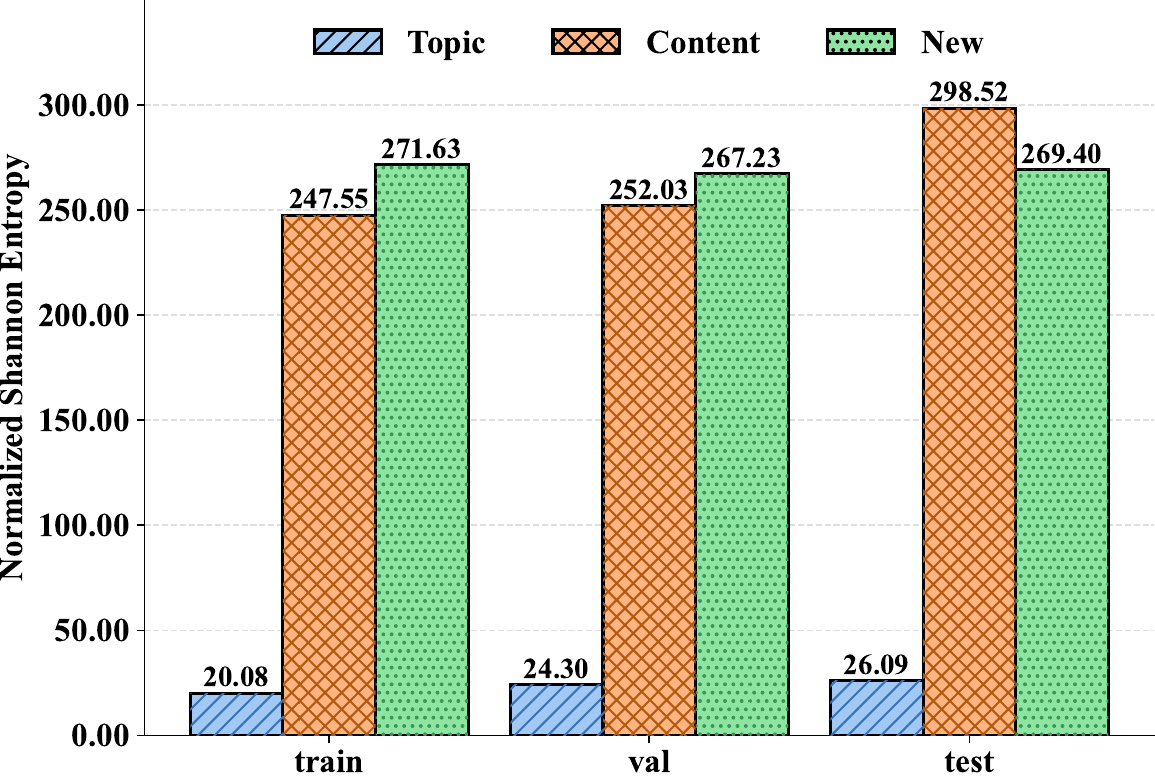}
    \caption{Weibo21 dataset Shannon Entropy}
    \label{fig:Chinese_Shannon.}
\end{subfigure}
\hfill
\begin{subfigure}[t]{0.485\linewidth}
    \centering
    \includegraphics[width=\linewidth]{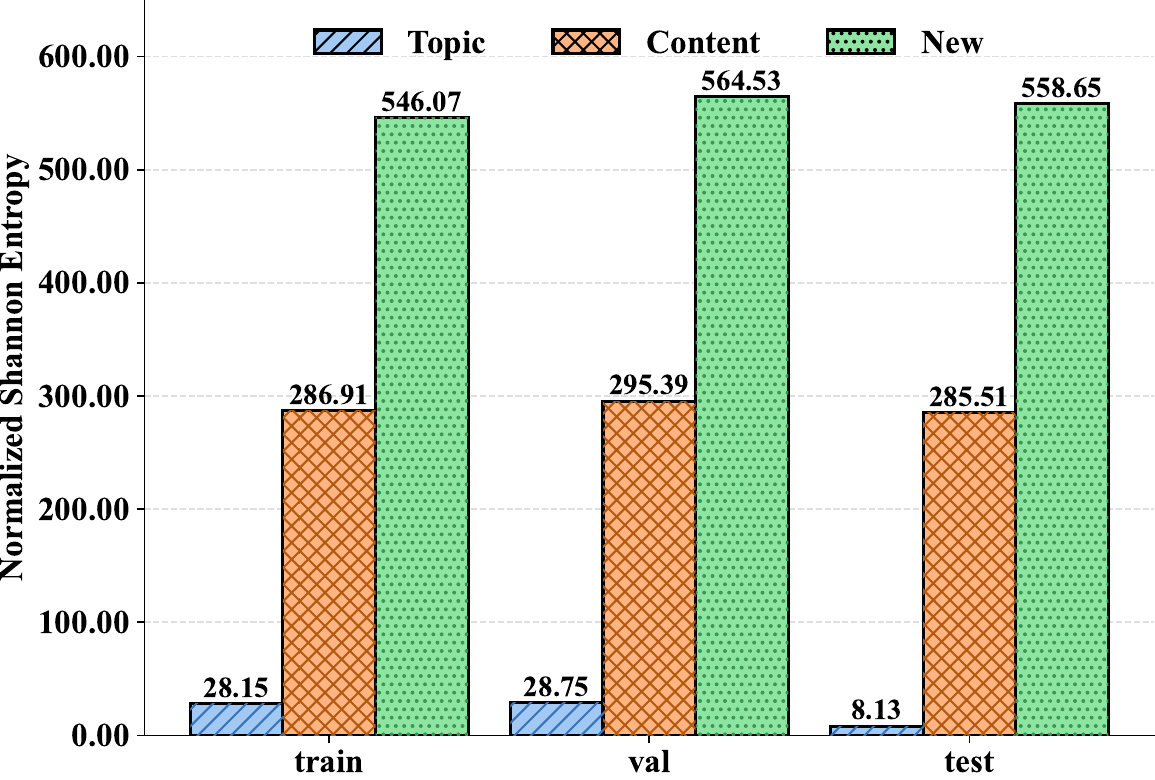}
    \caption{GossipCop dataset Shannon Entropy}
    \label{fig:English_Shannon.}
\end{subfigure}

\caption{Similarity and Shannon entropy analysis on Weibo21 and GossipCop datasets.}
\label{Figure 4-2}
\end{figure*}

\subsection{News Topic-Content Extraction}
Given that the topic of a news article best encapsulates the overall meaning, and the main content centers on the core event—thus minimizing stylistic influence—our method employs LLMs to extract both the topic and main content of news texts. The detailed extraction process is illustrated in the left half of Figure~\ref{Prompt}. Prompt semantics are kept consistent across Chinese and English, with translations performed as needed. During extraction, we assess both the semantic similarity and information density between the extracted text and the original news to evaluate relevance and mitigate the risk of LLM hallucinations. Detailed experimental results are presented in Section~\ref{extraction}. Owing to differences in grammatical structure and information density between Chinese and English, the similarity scores for Weibo21 (Chinese) are typically lower than those for GossipCop (English). Accordingly, we require that the cosine similarity between extracted English news topic-content and the original news exceeds 0.9, while for extracted Chinese news topic-content the threshold is set at 0.8 according to the distribution of most similarity scores. If these criteria are not met, the LLM is prompted to regenerate the extraction use advanced LLM, such as Chatgpt-4o~\cite{openai_gpt4o_mini_blog} and DeepSeek-R1~\cite{deepseekai2025}. In addition, after extraction, we further evaluate the information content of the results using Shannon entropy.

\subsection{Large Model Extraction Metrics}
\label{extraction}

By evaluating both cosine similarity and Shannon entropy, we ensure that the extracted content is both faithful to the original news and sufficiently informative.

First, we input both the generated text and the real text into the corresponding Text Encoder, and assess their semantic similarity using cosine similarity. This approach helps constrain potential hallucinations from the LLM during content extraction. As shown in the top panels of Figure~\ref{Figure 4-2}, the cosine similarity between the extracted topic-content, and the original news serves as a constraint during the extraction process—exceeding 0.8 for the Weibo21 dataset and 0.9 for the GossipCop dataset. If the similarity does not reach the specified threshold, the extraction process is repeated until the requirement is satisfied. Finally, across all data splits, the average cosine similarity between the extracted topic-content and the original news consistently exceeds 0.9 in two datasets. These results indicate that the extracted topics and content are highly semantically aligned with the original news, demonstrating that the LLM fulfills our requirements with minimal hallucination.

Next, we calculate the Shannon entropy for both the generated and real texts to evaluate whether the generated content retains sufficient information~\cite{cao2025less}. A significantly lower Shannon entropy in the generated text compared to the real text may indicate overly simplified content and possible hallucination. The bottom panels of Figure~\ref{Figure 4-2} present the average entropy values per information piece in the datasets. Here, notable differences emerge between Weibo21 and GoosipCop datasets. The original English news is longer and has a higher information density, yet after extraction, the information density drops to about half that of the original text. In contrast, the original Chinese news is shorter, and due to the inherently high information density of Chinese, the extracted topic-content maintains a similar information density as the original. Overall, the high semantic similarity and reasonable information density of the extracted content indicate that the LLM-based extraction process is both precise and reliable, with low risk of content hallucination across both Weibo21 and GossipCop datasets.

Through these two indicators, we screened and evaluated the content output by LLMs, and finally reduced the illusion of LLMs in the task of extracting news topic-content as much as possible.

\subsection{Commonsense Reasoning and Judgment}
Commonsense reasoning leverages the inherent knowledge base of LLM to identify contradictions within news content from a commonsense perspective, aligning more closely with general human cognition. This component follows the approach proposed in ARG~\cite{hu2024bad}, with both the Weibo21 and GossipCop datasets' commonsense reasoning and judgment modules implemented using ChatGPT3.5-turbo. Prompt semantics are consistent across both languages, with translations as needed. The corresponding workflow is illustrated in the right half of Figure~\ref{Prompt}.
\begin{figure*}[t]
\centering
\begin{subfigure}[t]{0.48\linewidth}
    \centering
    \includegraphics[width=\linewidth]{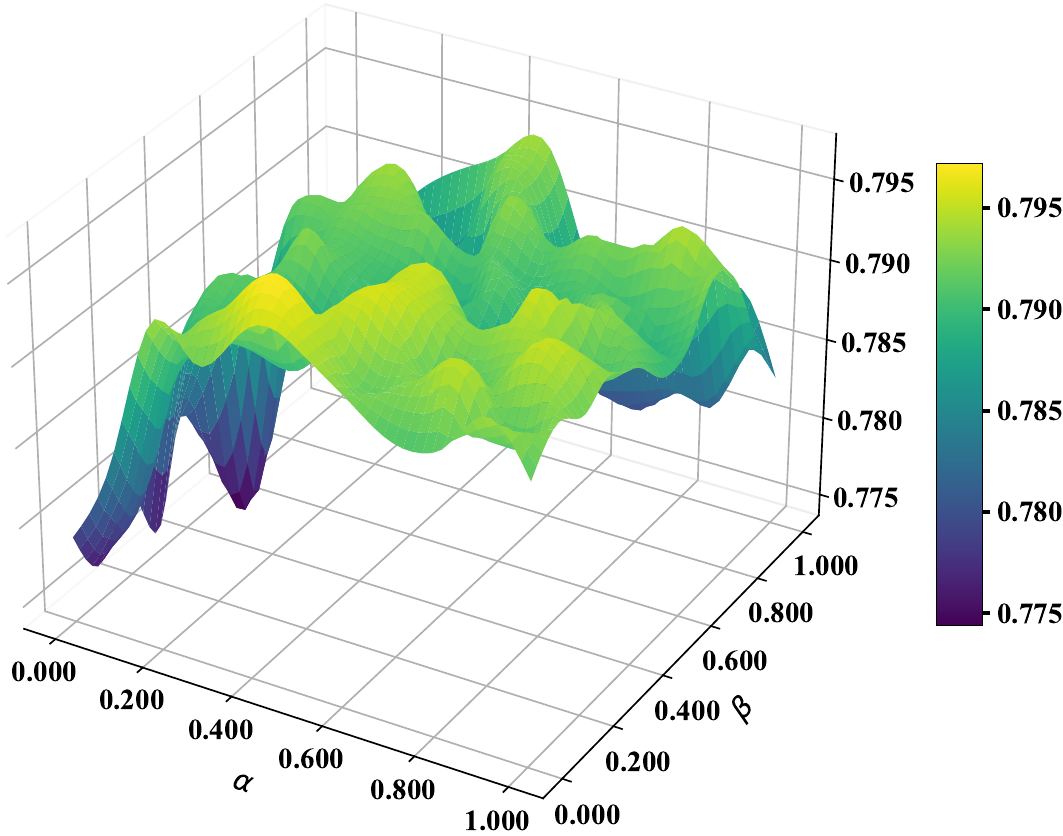}
    \caption{the macF1 of Weibo21 dataset}
    \label{fig:zh_f1mac}
\end{subfigure}%
\hspace{0.03\linewidth}
\begin{subfigure}[t]{0.48\linewidth}
    \centering
    \includegraphics[width=\linewidth]{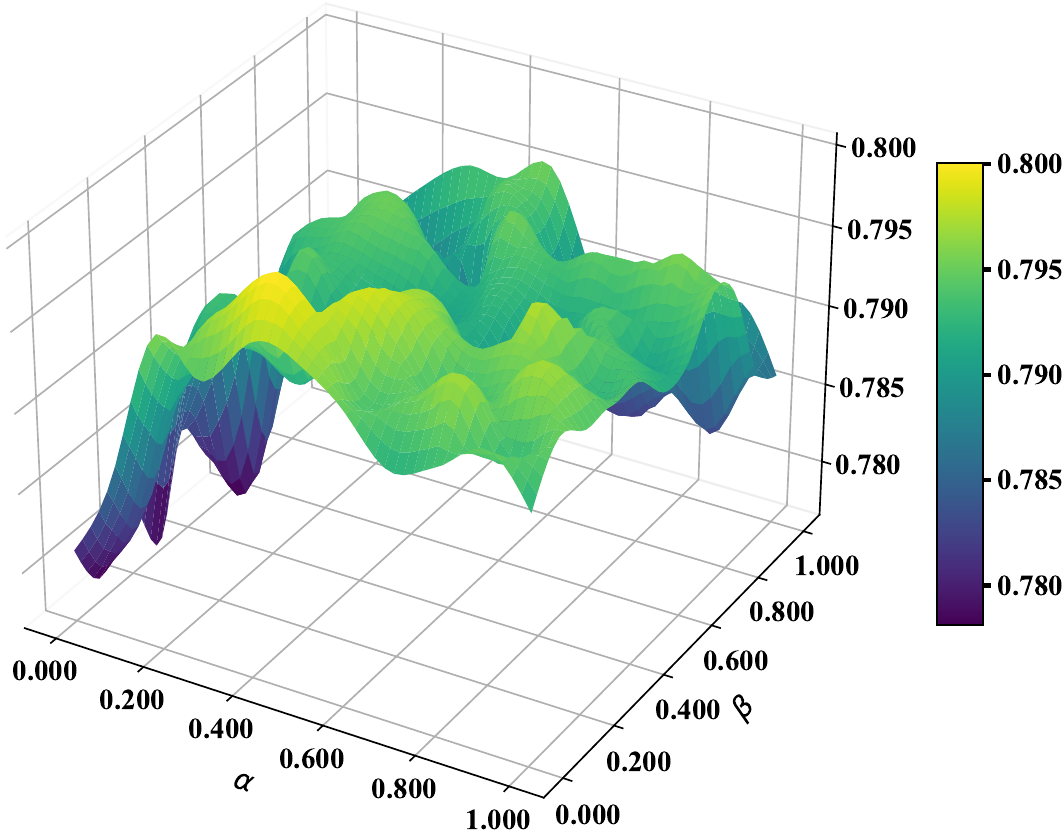}
    \caption{the Acc. of Weibo21 dataset}
    \label{fig:zh_acc}
\end{subfigure}
\end{figure*}

\begin{figure*}[t]
\ContinuedFloat
\centering
\begin{subfigure}[t]{0.48\linewidth}
    \centering
    \includegraphics[width=\linewidth]{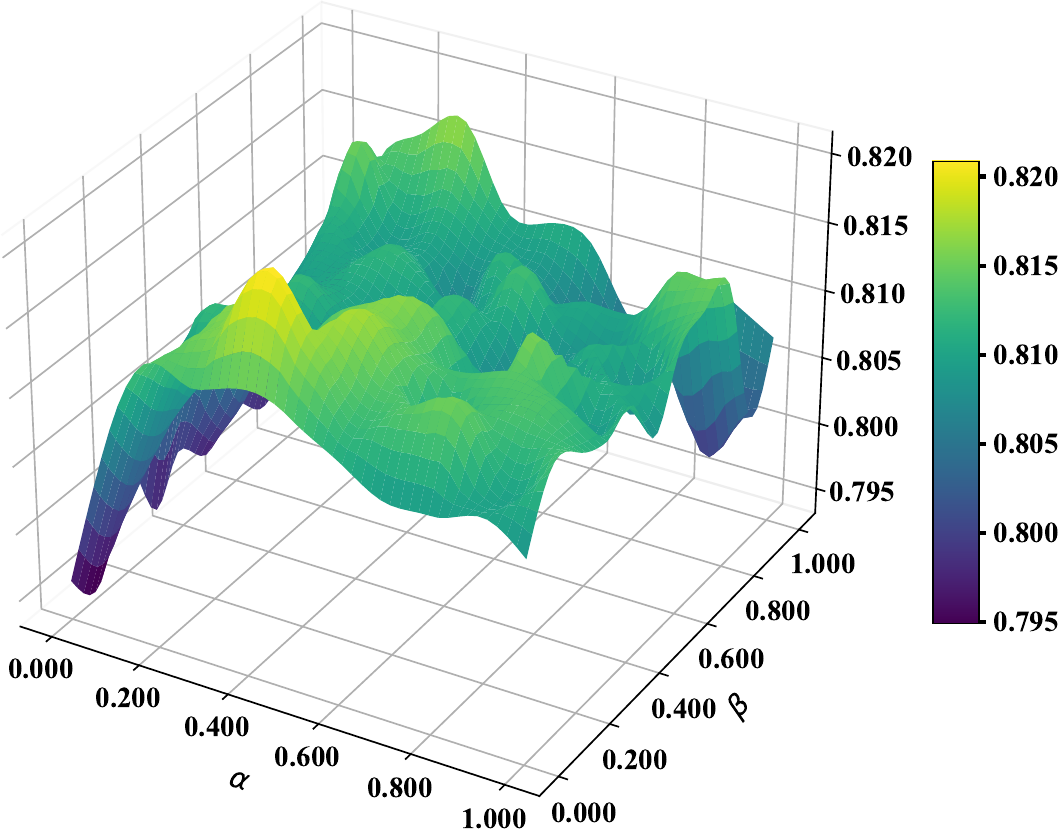}
    \caption{the $\text{F1}_\text{real}$ of Weibo21 dataset}
    \label{fig:zh_f1real}
\end{subfigure}%
\hspace{0.03\linewidth}
\begin{subfigure}[t]{0.48\linewidth}
    \centering
    \includegraphics[width=\linewidth]{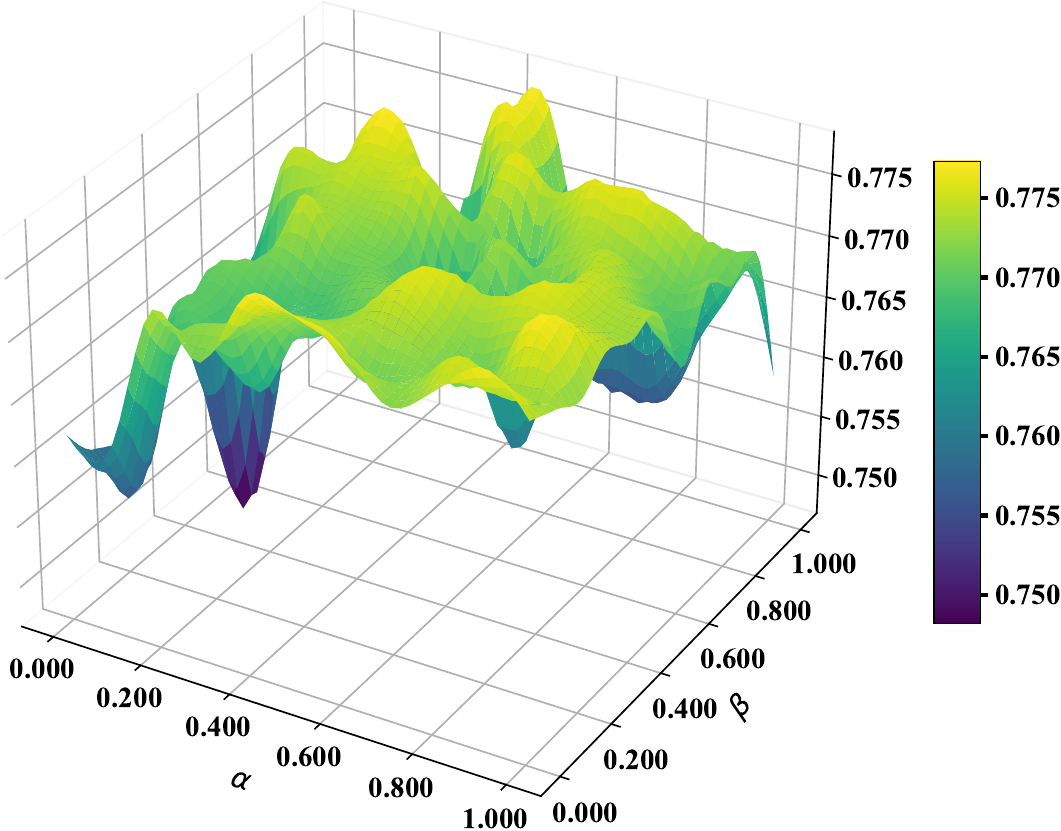}
    \caption{the $\text{F1}_\text{fake}$ of Weibo21 dataset}
    \label{fig:zh_f1fake}
\end{subfigure}
\end{figure*}

\begin{figure*}[t]
\ContinuedFloat
\centering
\begin{subfigure}[t]{0.48\linewidth}
    \centering
    \includegraphics[width=\linewidth]{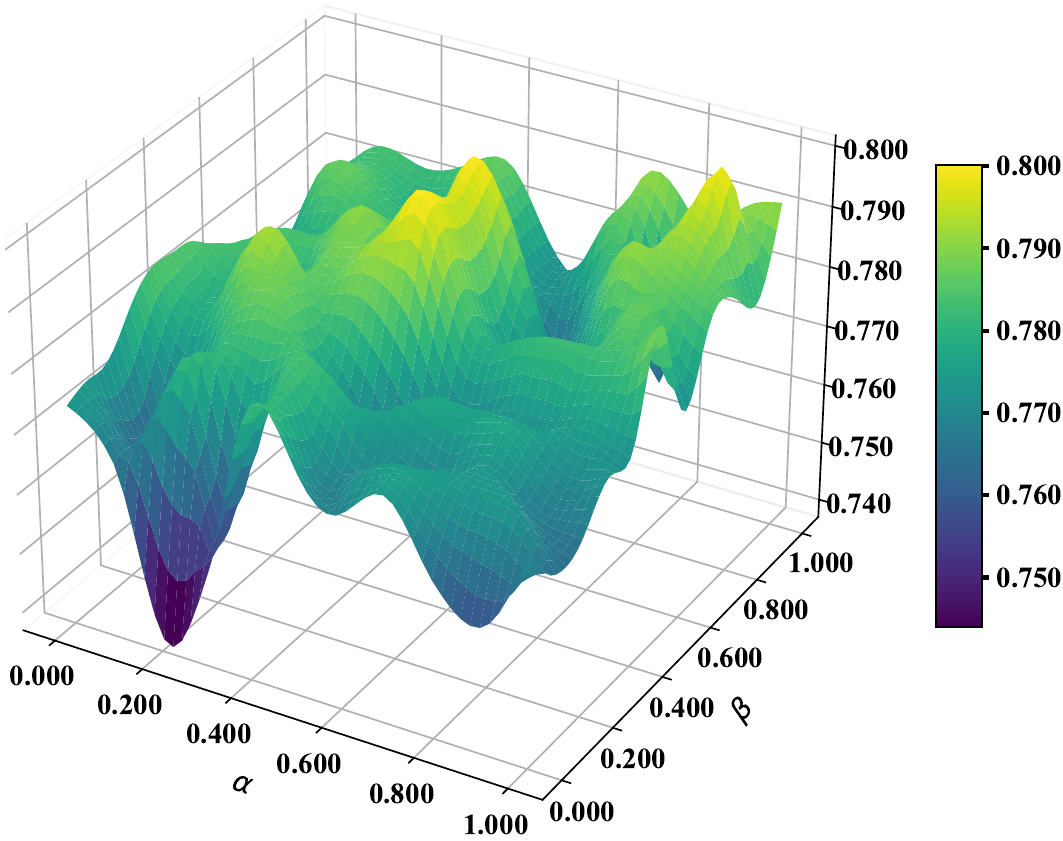}
    \caption{the macF1 of GossipCop dataset}
    \label{fig:en_f1mac}
\end{subfigure}%
\hspace{0.03\linewidth}
\begin{subfigure}[t]{0.48\linewidth}
    \centering
    \includegraphics[width=\linewidth]{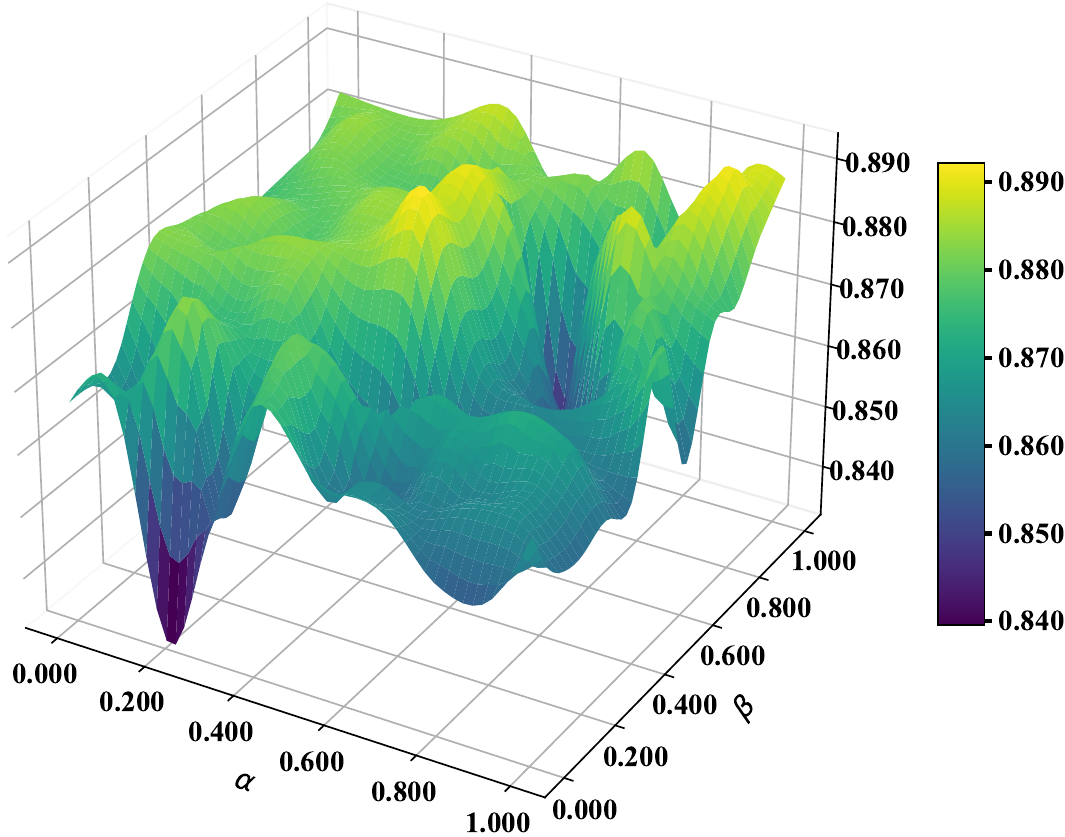}
    \caption{the Acc. of GossipCop dataset}
    \label{fig:en_acc}
\end{subfigure}

\ContinuedFloat
\centering
\begin{subfigure}[t]{0.48\linewidth}
    \centering
    \includegraphics[width=\linewidth]{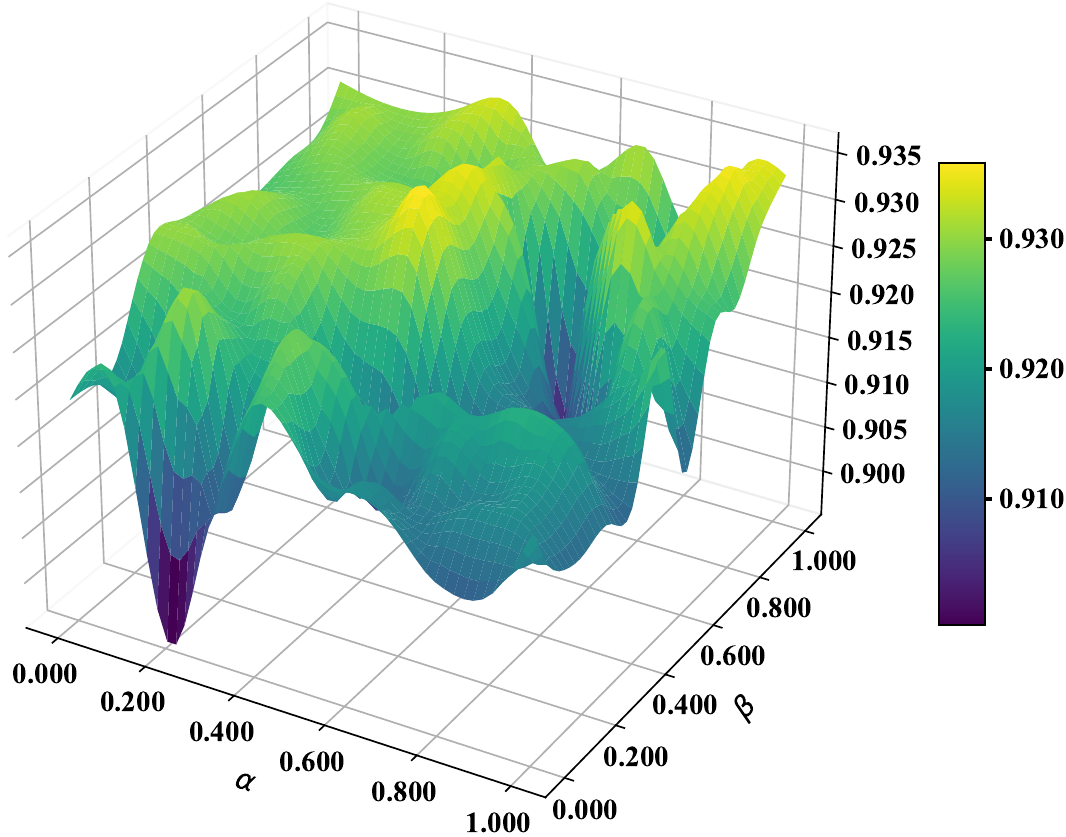}
    \caption{the $\text{F1}_\text{real}$ of GossipCop dataset}
    \label{fig:en_f1real}
\end{subfigure}%
\hspace{0.03\linewidth}
\begin{subfigure}[t]{0.48\linewidth}
    \centering
    \includegraphics[width=\linewidth]{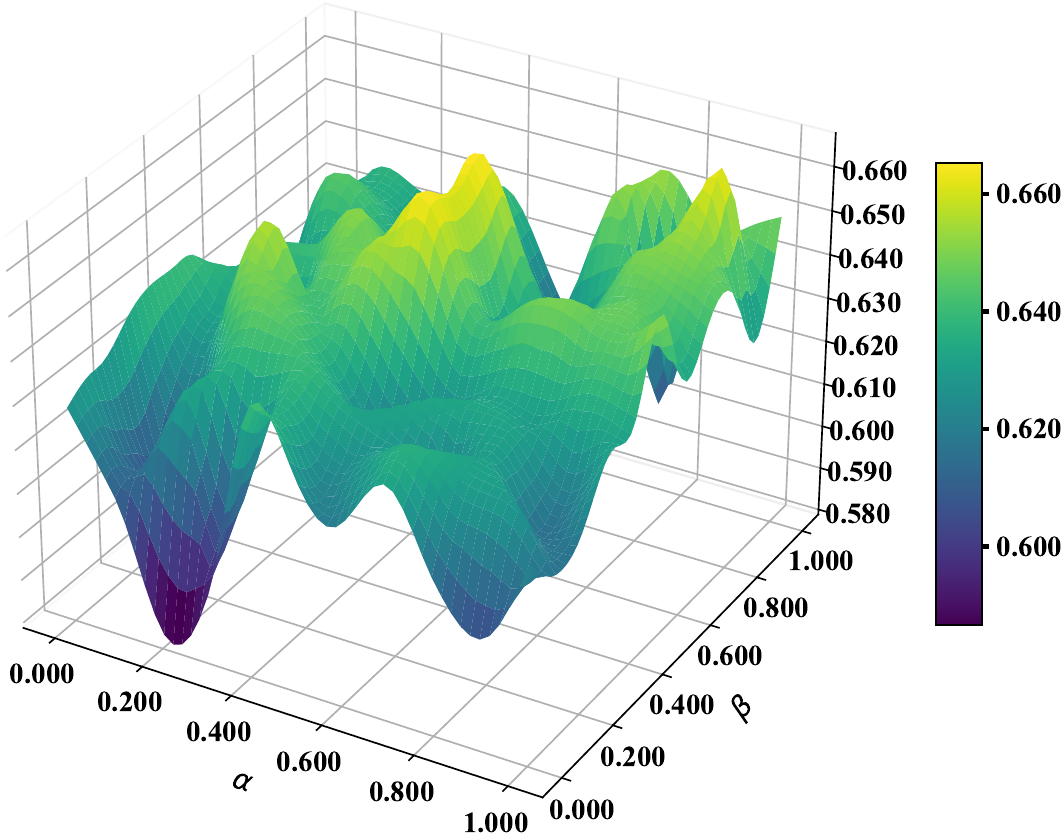}
    \caption{the $\text{F1}_\text{fake}$ of GossipCop dataset}
    \label{fig:en_f1fake}
\end{subfigure}
\caption{Sensitivity analysis of \factguard model in Weibo21 and GossipCop datasets across four evaluation metrics: $\text{macF1}$, $\text{Accuracy}$, $\text{F1}_\text{real}$, and $\text{F1}_\text{fake}$.}
\label{sensitivity}
\end{figure*}
\section{Experimental Setup}
\label{app_sec:experimental_setup}

\begin{figure*}[t]  
\centering


\begin{subfigure}[t]{0.485\linewidth}
    \centering
    \includegraphics[width=\linewidth]{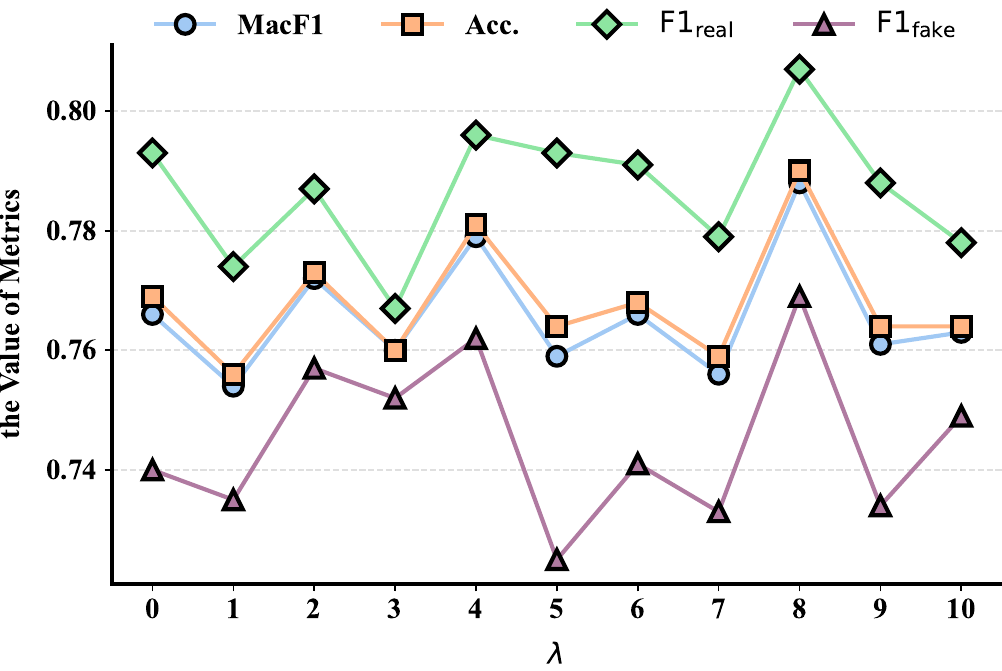}
    \caption{the value of Weibo21 dataset distillation result}
\end{subfigure}
\hfill
\begin{subfigure}[t]{0.485\linewidth}
    \centering
    \includegraphics[width=\linewidth]{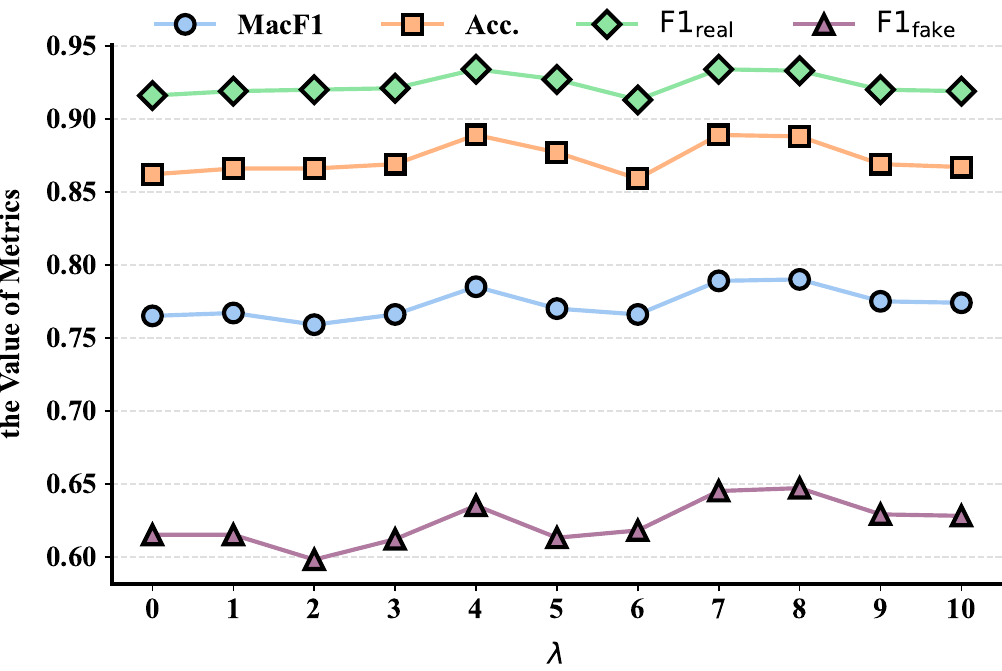}
    \caption{the value of GossipCop dataset distillation result}
\end{subfigure}

\caption{Sensitivity analysis of \factguard -D model in Weibo21 and GossipCop datasets across four evaluation metrics: $\text{macF1}$, $\text{Acc.}$, $\text{F1}_\text{real}$, and $\text{F1}_\text{fake}$.}
\label{lambda}
\end{figure*}

\subsection{Dataset}
We employ the Weibo21 (Chinese)~\citep{nan2021mdfend} and GossipCop (English)~\citep{shu2020fakenewsnet} for evaluation. Both datasets are preprocessed by deduplication and temporal splitting, following established practices~\citep{zhu2022generalizing,mu2023s,hu2024bad}, to mitigate the risk of data leakage and prevent overestimation of SLM performance. In addition, we also utilize the commonsense rationales from \cite{hu2024bad}.The statistical summaries of these datasets are provided in Table~\ref{Table 4-1}.

\begin{table}[htbp]
\centering
\small
\renewcommand{\arraystretch}{1.1}
\setlength{\tabcolsep}{2mm}
\begin{tabular}{c ccc ccc}
\toprule
\multirow{2}{*}{\textbf{\#}} & \multicolumn{3}{c}{\textbf{Weibo21}} & \multicolumn{3}{c}{\textbf{GossipCop}} \\
\cmidrule(lr){2-4} \cmidrule(lr){5-7}
& \textbf{Train} & \textbf{Val} & \textbf{Test} & \textbf{Train} & \textbf{Val} & \textbf{Test} \\
\midrule
Real & 2,331 & 1,172 & 1,137 & 2,878 & 1,030 & 1,024 \\
Fake & 2,873 & 779 & 814 & 1,006 & 244 & 234 \\
\cdashline{1-7} 
Total & 5,204 & 1,951 & 1,951 & 3,884 & 1,274 & 1,258 \\
\bottomrule
\end{tabular}
\setlength{\abovecaptionskip}{5pt}
\setlength{\belowcaptionskip}{-10pt}

\caption{Statistics of the number of real and fake samples in Weibo21 and GossipCop datasets.}
\label{Table 4-1}
\end{table}

\subsection{Baselines}
\label{sec:baselines}

Recent advances in early fake news detection predominantly leverage LLMs and SLMs which only uses news content. In this study, we select 16 representative methods and categorize them into four groups: Group 1 comprises LLM-only methods; Group 2 consists of SLM-only methods; Group 3 includes LLM-SLM methods; and Group 4 focuses on knowledge distillation.

\subsubsection{G1: LLM-only.}
These methods directly employ prompt engineering and multi-agent frameworks with LLMs for fake news detection.
\begin{enumerate}
    \item \textbf{GPT-3.5-turbo}~\cite{openai2023gpt35}: Employed in conjunction with few-shot learning for Weibo21 dataset and few-shot CoT for GossipCop dataset.
    \item \textbf{GPT-4o-mini}~\cite{openai_gpt4o_mini_blog}: Direct use of GPT-4o-mini as an LLM detector for fake news detection.
    \item \textbf{ChatEval-o}~\cite{chan2024chateval}: Utilizes the one-by-one strategy for fake news migration within the multi-agent debate framework ChatEval.
    \item \textbf{ChatEval-s}~\cite{chan2024chateval}: Applies the Simultaneous-Talk strategy in the ChatEval multi-agent debate framework to tackle the fake news migration task.
\end{enumerate}

\subsubsection{G2: SLM-only.}
SLMs, such as BERT or RoBERTa, generally perform well on fake news detection tasks. This category includes:
\begin{enumerate}
    \item \textbf{BERT}~\cite{devlin2019bert}: Fake news detection using a fine-tuned vanilla BERT-base model.
    \item \textbf{RoBERTa}~\cite{liu2019roberta}: Employs RoBERTa as the text encoder, incorporating linear attention for enhanced fake news detection.
    \item \textbf{EANN}~\cite{wang2018eann}: Employs auxiliary adversarial training to isolate and reduce event-related features, using the publication year as an auxiliary label.
    \item \textbf{Publisher-Emo}~\cite{zhang2021mining}: Integrates emotional attributes with textual features for fake news detection.
    \item \textbf{ENDEFA}~\cite{zhu2022generalizing}: Utilizes causal learning to eliminate entity bias, enhancing generalization under distribution shifts. 
\end{enumerate}

\subsubsection{G3: LLM-SLM.}
LLMs alone often do not perform optimally. Therefore, several approaches combine LLM and SLM to enhance performance.
\begin{enumerate}
    \item \textbf{BERT+Rationale}~\cite{hu2024bad}: Combines features from both the news and rationale encoders, feeding them into an MLP for prediction.
    \item \textbf{SuperICL}~\cite{zhong2023can}: Uses an SLM plug-in to enhance the in-context learning capabilities of the LLM by incorporating predictions and confidence levels for each test sample into the prompt.
    \item \textbf{BERT+GenFEND}~\cite{nan2024let}: Combines features from both the news and LLMs' comments based on BERT and qwen3-235b-a22b-instruct-2507 \footnote{\url{https://huggingface.co/Qwen/Qwen3-235B-A22B-Instruct-2507}}~\cite{qwen3technicalreport}.
    \item \textbf{RoBERTa+GenFEND}~\cite{nan2024let}: Combines features from both the news and LLMs' comments based on Roberta and qwen3-235b-a22b-instruct-2507.
    \item \textbf{TED}~\cite{liu2025truth}: A multi-agent system leveraging LLM-driven structured debates to enhance both interpretability and accuracy.
\end{enumerate}

\subsubsection{G4: Distilled Model.}
Distilled models are particularly suitable for resource-constrained and cold-start scenarios.
\begin{enumerate}
    \item \textbf{ARG-D}~\cite{hu2024bad}: A rationale-free version of ARG created via distillation, designed for cost-sensitive applications that do not require LLM queries.
\end{enumerate}

\subsection{Metrics}
\label{sec:evaluation}

This study employs four evaluation metrics: Accuracy ($\mathrm{Acc.}$), $\mathrm{F1}_\mathrm{real}$, $\mathrm{F1}_\mathrm{fake}$, and Macro-F1 ($\mathrm{macF1}$), to comprehensively assess the performance of the \factguard model.

\paragraph{$\text{Acc.}$} measures the proportion of all news samples (both real and fake) that are correctly classified by the model:
\begin{equation}
\text{Acc.} = \frac{\text{TP + TN}}{\text{TP + TN + FP + FN}},
\end{equation}
where $\text{TP}$ (True Positive) is the number of fake news articles correctly identified, $\text{TN}$ (True Negative) is the number of real news articles correctly identified, $\text{FP}$ (False Positive) is the number of real news articles incorrectly classified as fake, and $\text{FN}$ (False Negative) is the number of fake news articles incorrectly classified as real. Together, these four quantities constitute the confusion matrix, providing a comprehensive view of the model’s classification performance.

\paragraph{$\text{F1}_\text{real}$} is the F1 score for the ``real news'' category, reflecting the model’s performance in identifying real news:
\begin{equation}
{\text{F1}_\text{real}} = 2 \times \frac{\text{Precision}_\text{real} \times {\text{Recall}_\text{real}}}{\text{Precision}_\text{real} + \text{Recall}_\text{real}},
\end{equation}
where $\text{Precision}_\text{real}$ denotes the proportion of news predicted as real that is actually real, and $\text{Recall}_\text{real}$ represents the proportion of true real news that is correctly identified.

\paragraph{$\text{F1}_\text{fake}$} is the F1 score for the ``fake news'' category, emphasizing the model's detection ability for fake news:
\begin{equation}
\text{F1}_\text{fake} = 2 \times \frac{{\text{Precision}_\text{fake}} \times {\text{Recall}_\text{fake}}}{\text{Precision}_\text{fake} + {\text{Recall}_\text{fake}}},
\end{equation}
where $\text{Precision}_\text{fake}$ denotes the proportion of news predicted as fake that is actually fake, and $\text{Recall}_\text{fake}$ indicates the proportion of true fake news that is successfully detected.

\paragraph{$\text{macF1}$} is the average $\text{F1}$ score across both categories, providing an overall measure of recognition capability:
\begin{equation}
\text{macF1} = \frac{{\text{F1}_\text{fake}} + {\text{F1}_\text{real}}}{2}.
\end{equation}

\begin{table*}[ht]
\centering
\small
{
\begin{tabular}{llcccccccc}
\toprule

& \multirow{2}{*}{\textbf{Text Encoder}} & \multicolumn{4}{c}{\textbf{Weibo21}} & \multicolumn{4}{c}{\textbf{GossipCop}} \\
\cmidrule(lr){3-6} \cmidrule(lr){7-10}
& & macF1 & Acc. & F1$_\text{real}$ & F1$_\text{fake}$ & macF1 & Acc. & F1$_\text{real}$ & F1$_\text{fake}$ \\
\midrule

\multirow{1}{*} 
& Bert & \textbf{0.801} & \textbf{0.804} & \textbf{0.824} & \textbf{0.777} & 0.784 & 0.879 & 0.927 & 0.642 \\ 
& RoBERTa & 0.764 & 0.765 & 0.784 & 0.745 & \textbf{0.805} & \textbf{0.892} & \textbf{0.935} & \textbf{0.675} \\

\bottomrule
\end{tabular}
}  
\caption{\footnotesize the performance of BERT and RoBERTa as \factguard 's encoder.}
\label{Table 4-4}
\end{table*}

\begin{figure*}[t]
\centering
\includegraphics[width=1.0\textwidth]{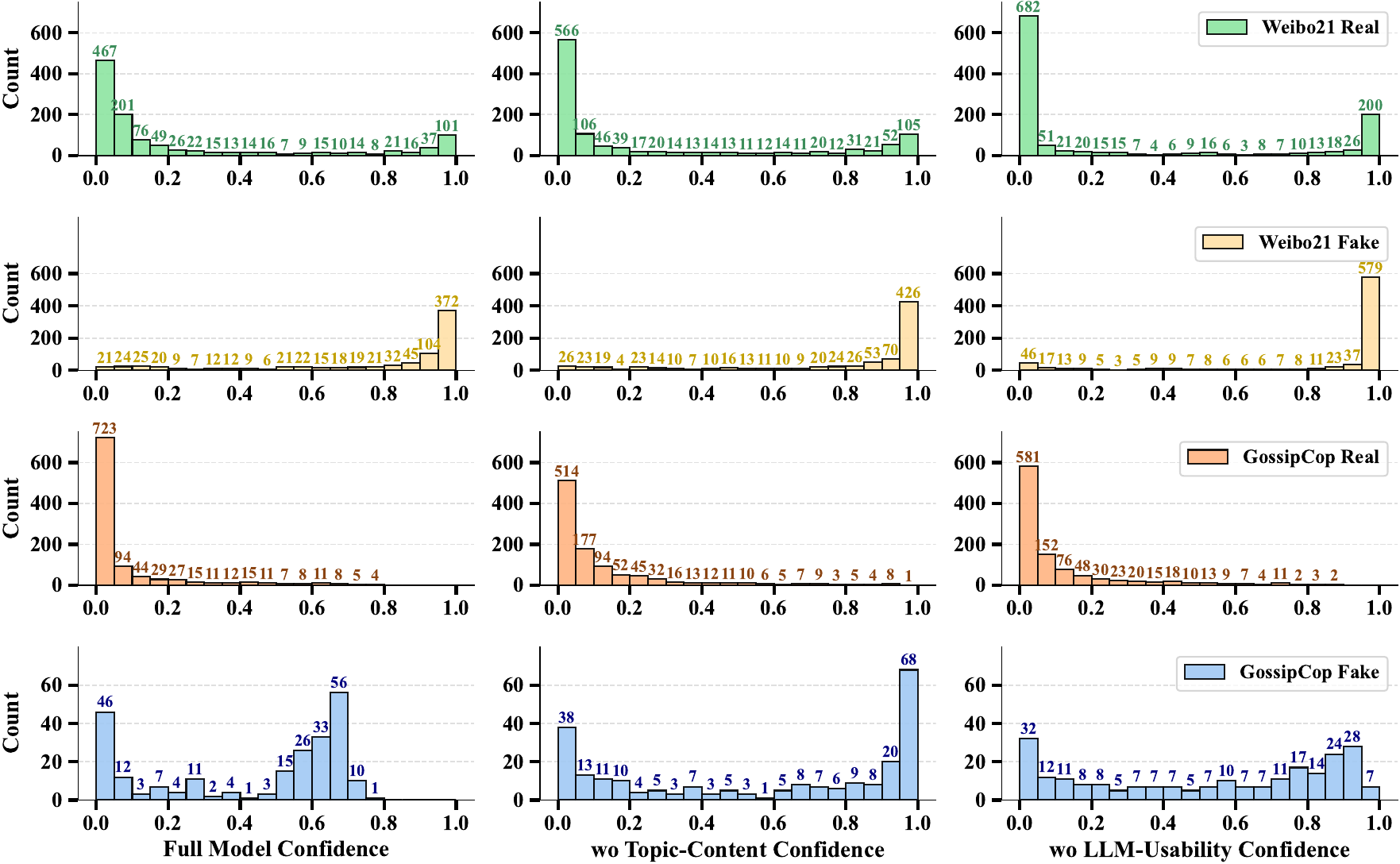}
\caption{
Confidence distributions of real and fake samples in GossipCop and Weibo21 dataset before and after with style debiasing and llm usability judgment in \factguard.
}
\label{confidence}
\end{figure*}

\section{Sensitivity Study}
\label{sec:sensitivity_study}

This section presents parameter sensitivity experiments for two loss function hyperparameters as well as the choice of text encoder of \factguard and \factguard -D to identify the optimal model configuration.

\subsection{Loss Hyperparameters}
The loss function of \factguard involves two key hyperparameters, $\alpha$ and $\beta$. We employ a grid search to determine their optimal values. As illustrated in Figure~\ref{sensitivity}, we evaluate four performance metrics (i.e. $\text{Acc.}$, $\text{F1}_\text{real}$, $\text{F1}_\text{fake}$ and $\text{macF1}$) on both Weibo21 and GossipCop datasets under various hyperparameter settings. In the plots, the x-axis denotes $\alpha$, the y-axis denotes $\beta$, and the z-axis represents the value of the corresponding metric. Both $\alpha$ and $\beta$ are discretized into 11 grid points from 0 to 10. Figure~\ref{sensitivity} displays the results of this grid search.

The results indicate that \factguard achieves optimal performance on the Weibo21 dataset when $\alpha=0.4$ and $\beta$ is within the range (0.1, 0.2). For the GossipCop dataset, the best results of \factguard are observed when $\alpha=0.5$ and $\beta$ falls within the range (0.5, 0.6). Based on these findings, we further conduct a random search for $\beta$ and ultimately determine the optimal hyperparameters of \factguard to be $\alpha=0.40$, $\beta=0.16$ in Chinese \factguard, and $\alpha=0.50$, $\beta=0.58$ in English \factguard.

The loss function of \factguard-D involves a key hyperparameter, the distillation coefficient $\lambda$. As shown in Figure~\ref{lambda}, we evaluate eight performance metrics on both Weibo21 and GossipCop datasets under different hyperparameter settings. In the plots, the x-axis represents $\lambda$, while the y-axis indicates the value of the corresponding metric. The value of $\lambda$ is discretized into 11 points ranging from 0 to 10. Figure~\ref{lambda} presents the results of this sequential hyperparameter search. The results demonstrate that the optimal performance on both the Weibo21 and GossipCop datasets is achieved when $\lambda=8$.

\subsection{Text Encoder}
Due to differences in language characteristics between Chinese and English, the choice of text encoder can significantly affect \factguard performance. The comparative results for BERT and RoBERTa are summarized in Table~\ref{Table 4-4}. For Chinese \factguard, BERT is sourced from Google’s bert-base-chinese\footnote{\url{https://huggingface.co/google-bert/bert-base-chinese}}~\cite{devlin2019bert} pretrained model, and RoBERTa from hfl’s chinese-roberta-wwm-ext\footnote{\url{https://huggingface.co/hfl/chinese-roberta-wwm-ext}}~\cite{cui-etal-2020-revisiting} pretrained model. For English \factguard, BERT is based on Google’s bert-base-uncased pretrained model\footnote{\url{https://huggingface.co/google-bert/bert-base-uncased}}, and RoBERTa from Facebook’s roberta-base\footnote{\url{https://huggingface.co/FacebookAI/roberta-base}}~\cite{liu2019roberta} pretrained model. Based on the experimental results, Google’s BERT is selected as the Chinese \factguard text encoder and Facebook’s RoBERTa is chosen for English \factguard.

\begin{figure*}[t]
\centering
\includegraphics[width=1.0\textwidth]{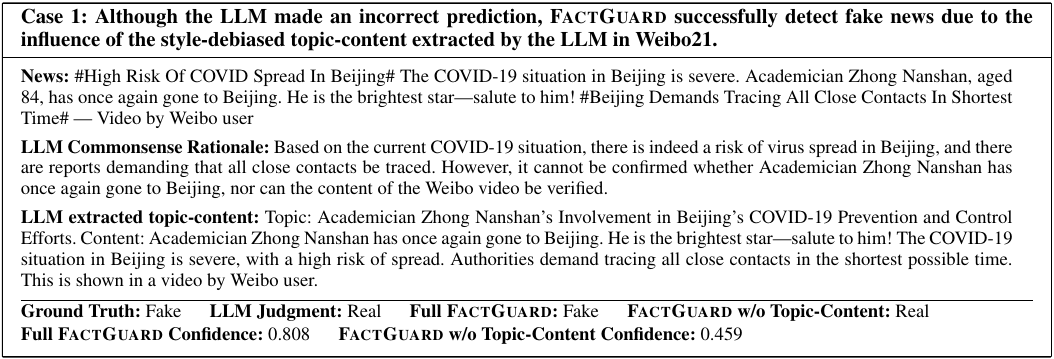}
\includegraphics[width=1.0\textwidth]{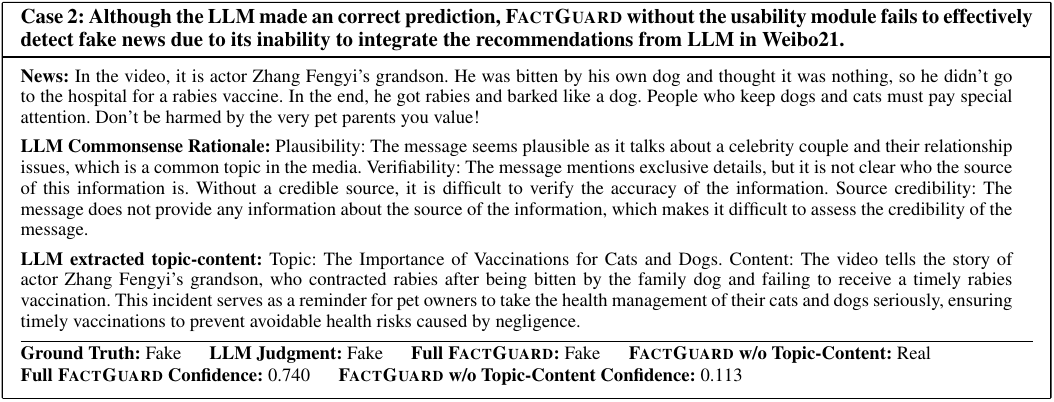}
\end{figure*}
\begin{figure*}[t]
\centering
\includegraphics[width=1.0\textwidth]{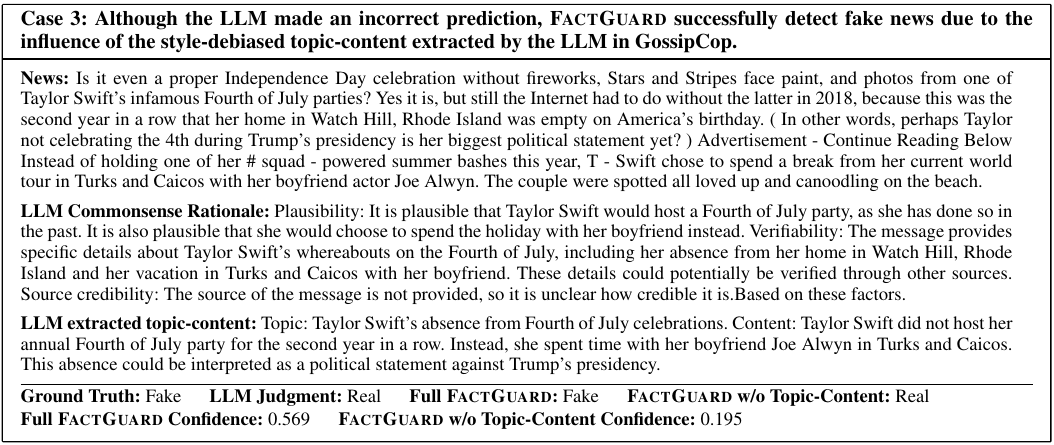}
\includegraphics[width=1.0\textwidth]{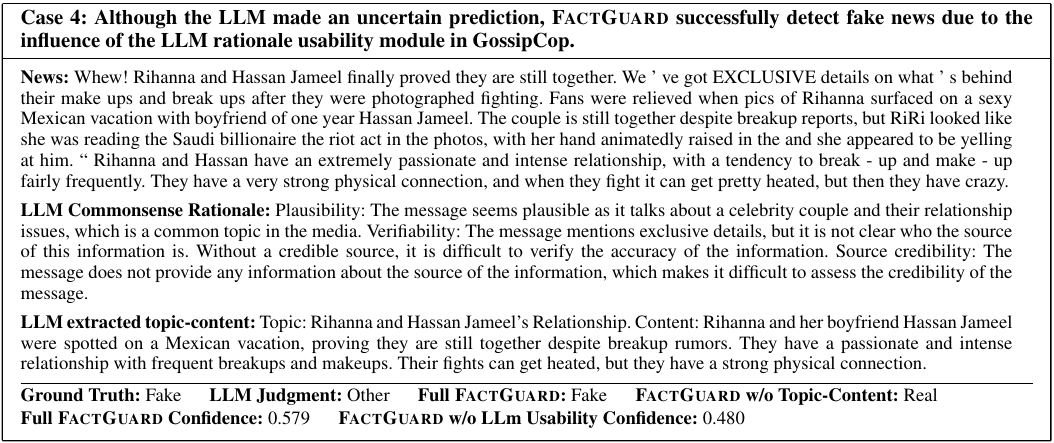}
\caption{
Case analysis: style-debiased topic-content and llm usability modules boost \factguard's fake news detection.
}
\label{case}
\end{figure*}

\subsection{Confidence Distribution Analysis}
Figure~\ref{confidence} depicts the confidence distributions of real and fake samples on the Weibo21 and GossipCop datasets under three model configurations: the full \factguard model, the ablation without topic-content features, and the ablation without LLM usability evaluator. This comparison allows us to disentangle the unique contributions of style debiasing and LLM-based factual judgment to model calibration and detection performance.

Across both datasets, we observe that the full \factguard model, which integrates both style debiasing (via topic-content disentanglement) and LLM usability evaluation, yields the most desirable and interpretable confidence distributions. For fake news samples, the full \factguard model markedly suppresses the occurrence of extremely high-confidence predictions (i.e., fake samples prediction probabilities from 0.8 to 1). This moderation of confidence is not merely a numerical adjustment; rather, it reflects the model’s enhanced ability to avoid overfitting to superficial stylistic artifacts and spurious correlations that are often present in fake news. Importantly, this reduction in overconfident judgments does not compromise overall detection performance; on the contrary, the model demonstrates improved effectiveness in identifying fake news, indicating that the learned decision boundary is more robust and generalizable. The resulting confidence scores for fake news become more evenly distributed, suggesting better calibration and a lower risk of making extreme, potentially erroneous decisions.

For fake news samples, when we ablate the topic-content module, thereby removing the style debiasing mechanism, a clear shift in the confidence distribution emerges. The model tends to revert to overconfident predictions for fake samples, with a noticeable increase in the number of samples assigned near-certain fake probabilities. This pattern indicates a renewed reliance on shallow stylistic or topical cues that are frequently entangled with fake news, leading to poorer generalization. The absence of topic-content disentanglement thus exposes the model to the risk of exploiting dataset-specific artifacts, underlining the critical role of style debiasing in mitigating such bias and promoting more reliable, fact-oriented detection.

Ablating the LLM usability component yields a similarly instructive outcome. Without the guidance of LLM-based factual judgment, the model again displays a tendency toward extreme confidence for fake news and less decisive recognition of real news. The removal of this module results in a less nuanced decision process: fake samples are concentrated at the high-confidence end, while real samples suffer from a reduction in high-confidence, correct predictions and a concomitant increase in low-confidence assignments. This highlights the importance of LLM usability in enforcing a fact-centric evaluation framework, which complements style debiasing by promoting decisions grounded in semantic and factual consistency rather than surface-level features.

For real news samples, these trends are mirrored but in the opposite direction. The full \factguard model is able to assign a higher proportion of real samples with strong confidence for being real(i.e., real samples prediction probabilities from 0 to 0.2), while the number of low-confidence assignments decreases. This shift reflects the model’s increased decisiveness and accuracy, which is diminished when either the style debiasing or LLM usability module is removed. Notably, the topic-content ablation leads to greater uncertainty and misclassification for real news, reaffirming the importance of style disentanglement for robust recognition of authentic information.

In summary, the experimental results across both Weibo21 and GossipCop datasets consistently demonstrate that the integration of topic-content style debiasing and LLM usability evaluation not only optimizes detection performance but also produces confidence distributions that are more interpretable and reliable. The topic-content module plays a pivotal role in mitigating stylistic and topical biases, thereby preventing overfitting and promoting generalization, while LLM usability injects essential factual discernment into the decision process. The full \factguard model, benefiting from both mechanisms, achieves superior robustness, calibration, and real-world applicability in automated news verification.

\subsection{Case Analysis}
Figure~\ref{case} presents a comprehensive analysis of the impact of style-debiased topic-content extraction and the usability module within LLMs in automated fake news detection. Across both GossipCop and Weibo21 datasets, these modules together significantly enhance the reliability and interpretability of detection results.

Cases 1 and 3 illustrate how style-debiased topic-content extraction helps distill the factual core of news reports, even when the LLM initial prediction is incorrect. By filtering out emotional or exaggerated language, this module enables \factguard to focus on verifiable information and more accurately identify falsehoods. For example, in Case 1 (Weibo21), although the LLM incorrectly predicts the news as real, the topic-content extraction isolates the uncertainty about whether Zhong Nanshan actually traveled to Beijing, reducing reliance on sensational narrative and leading \factguard to the correct ``fake'' classification with much higher confidence. In Case 3 (GossipCop), even after the LLM mislabels the news as real, the extraction process highlights the factual absence of Taylor Swift’s Fourth of July party, helping \factguard ignore the gossip and focus on checkable facts, thereby improving robustness and accuracy in fake news detection.

Cases 2 and 4 highlight the limitations of the model when the LLM usability module is absent, making systematic evaluation of news credibility and verifiability difficult. In Case 2 (Weibo21), although the LLM correctly predicts the news as fake, without the usability module, \factguard fails to effectively detect key fake news signals, such as the lack of source credibility and unclear attribution, resulting in very low confidence. Similarly, in Case 4 (GossipCop), when the LLM judgment is uncertain (``Other''), the removal of the usability module leads to the failure in identifying the absence of authoritative sources or verifiable evidence in the story about Rihanna and Hassan Jameel, lowering both confidence and interpretability. In both cases, the usability module is essential for integrating rationales about source reliability and verifiability, substantially boosting \factguard's confidence and final judgment.

Overall, the synergy between topic-content extraction and usability analysis modules allows \factguard to compensate for limitations in direct LLM predictions. By enhancing both accuracy and interpretability, this modular approach demonstrates clear practical value for large-scale, automatic fake news detection in real-world scenarios.

\end{document}